\documentclass[sigconf, nonacm]{acmart}
\usepackage[keeplastbox]{flushend}
\usepackage{subfig}

\def\BibTeX{{\rm B\kern-.05em{\sc i\kern-.025em b}\kern-.08em
    T\kern-.1667em\lower.7ex\hbox{E}\kern-.125emX}}
\usepackage{amsmath,amssymb,amsfonts}

\usepackage{amsmath}
\mathchardef\mhyphen="2D 
\usepackage{amsfonts}
\newcommand{\LNorm}[1]{\mathit{L#1\mhyphen Norm}}
\usepackage{xcolor}
\usepackage{listings}
\usepackage{graphicx}
\usepackage{amssymb}
\usepackage{pifont}
\newcommand{\cmark}{\ding{51}}%
\newcommand{\xmark}{\ding{55}}%
\usepackage{hyperref}
\usepackage[inline]{enumitem}

\usepackage{float}
\usepackage{textcomp}
\usepackage[linesnumbered, ruled]{algorithm2e}
\SetKwRepeat{Do}{do}{while}%
\usepackage{comment}
\usepackage{multicol}

\newsavebox{\tempboxa}
\newsavebox{\tempboxb}
\newsavebox{\tempboxc}
\usepackage{multirow}
\usepackage{booktabs}
\usepackage{xspace}

\usepackage{etoolbox}
\AtBeginEnvironment{align}{%
  \fontsize{9}{6}%
}

\newcommand{\ra}[1]{\renewcommand{\arraystretch}{#1}}

\definecolor{mGreen}{rgb}{0,0.6,0}
\definecolor{mGray}{rgb}{0.5,0.5,0.5}
\definecolor{mPurple}{rgb}{0.58,0,0.82}
\definecolor{backgroundColour}{rgb}{0.95,0.95,0.92}

\definecolor{fabulous}{rgb}{0.958, 0.188, 0.478}

\definecolor{BLU}{rgb}{0.01, 0.08, .9}

\newcommand*{\myfont}{\fontfamily{lmss}\selectfont}
\DeclareTextFontCommand{\HTfont}{\myfont}

\lstdefinestyle{CStyle}{
    backgroundcolor=\color{backgroundColour},   
    commentstyle=\color{mGreen},
    keywordstyle=\color{magenta},
    numberstyle=\tiny\color{mGray},
    stringstyle=\color{mPurple},
    basicstyle=\footnotesize,
    breakatwhitespace=false,         
    breaklines=true,                 
    captionpos=b,                    
    keepspaces=true,                 
    numbers=left,                    
    numbersep=5pt,                  
    showspaces=false,                
    showstringspaces=false,
    showtabs=false,                  
    tabsize=2,
    language=C,
}

\lstdefinestyle{COMP}{
    backgroundcolor=\color{white},   
    commentstyle=\color{mGreen},
    keywordstyle=\color{blue},
    numberstyle=\tiny\color{mGray},
    stringstyle=\color{mPurple},
    basicstyle=\footnotesize\ttfamily,
    breakatwhitespace=false,         
    breaklines=true,                 
    captionpos=b,                    
    keepspaces=true,                 
    numbers=left,                    
    numbersep=5pt,                  
    showspaces=false,                
    showstringspaces=false,
    showtabs=false,                  
    tabsize=2,
    language=C
}

\newcommand{\OURLCORE}{GOBO} 

\newcommand{\OURL}{\textit{\OURLCORE}\xspace} 

\begin{document}

\title{GOBO: Quantizing Attention-Based NLP Models \\
for Low Latency and Energy Efficient Inference}




\author{Ali Hadi Zadeh}
\affiliation{%
  \institution{University of Toronto }
}
\email{hadizade@ece.utoronto.ca}

\author{Isak Edo}
\affiliation{%
  \institution{University of Toronto }
}
\email{edoisak@ece.utoronto.ca}

\author{Omar Mohamed Awad}
\affiliation{%
  \institution{University of Toronto }
}
\email{awadomar@ece.utoronto.ca}

\author{Andreas Moshovos}
\affiliation{%
  \institution{University of Toronto }
}
\email{moshovos@ece.utoronto.ca}


\begin{abstract}
Attention-based models have demonstrated remarkable success in various natural language understanding tasks. However, efficient execution remains a challenge for these models which are memory-bound due to their massive number of parameters. We present \OURL, a model quantization technique that compresses the vast majority (typically 99.9\%) of the 32-bit floating-point parameters of state-of-the-art BERT models and their variants to 3 bits while maintaining their accuracy. Unlike other quantization methods, \OURL does not require fine-tuning nor retraining to compensate for the quantization error. We present two practical hardware applications of \OURL. In the first \OURL reduces memory storage and traffic and as a result inference latency and energy consumption. This \OURL memory compression mechanism is plug-in compatible with many architectures; we demonstrate it with the TPU, Eyeriss, and an architecture using Tensor Cores-like units. Second, we present a co-designed hardware architecture that also reduces computation. Uniquely, the \OURL architecture maintains most of the weights in 3b even during computation, a property that: 
\begin{enumerate*}[label=(\roman*)]
  \item makes the processing elements area efficient, allowing us to pack more compute power per unit area,
  \item replaces most multiply-accumulations with additions, and
  \item reduces the off-chip traffic by amplifying on-chip memory capacity.
\end{enumerate*}
\end{abstract}

\settopmatter{printfolios=true}
\maketitle





      

\begingroup
\renewcommand\thefootnote{}\footnote{\noindent

\raggedright Accepted at the 53rd IEEE/ACM International Symposium on Microarchitecture\textsuperscript{\textregistered}--- MICRO 2020

}\addtocounter{footnote}{-1}\endgroup

\section{Introduction}
Modern computing platforms, from servers to mobile and embedded devices, are energy constrained. Techniques for improving their energy efficiency can yield several benefits. They can reduce the energy footprint of data centers and as a result operating costs and environmental impact. They can increase uptime for mobile devices, and they can boost the capability of all systems by allowing them to perform more computations per unit of time. Improving energy efficiency is imperative for deep learning workloads as they are particularly compute and memory intensive. More so, given that the trend has been towards neural network models that require more computations and memory.

Memory accesses, be it off- or on-chip, account for a significant fraction of overall energy consumption when executing neural models. 
{Memory footprint, bandwidth and energy limitations are most acute for attention-based models in language understanding tasks. Among them, the BERT family of natural language models~\cite{BERT} are today the benchmark delivering best-of-class accuracy. Their footprint, accesses, and execution time are dominated by the parameters~(weights) of their numerous attention layers. The largest and most accurate among those models, BERT-Large has a footprint of 1.12GB, while BERT-base sacrifices some accuracy to reduce footprint to 326MB. These original BERT models are particularly expensive to train and use 32b floating-point parameters even for inference. Training BERT-Large on 16 Cloud TPUs (64 TPU chips) takes 4 days~\cite{BERT}. For this reason, in typical applications we start with a pre-trained version which we then refine to a target task. Refinement time varies per task but usually takes hours to a few days for BERT-Large on an RTX~2080~Ti GPU.  Several architecture modifications and quantization methods have been proposed to reduce the cost of BERT models~\cite{sanh2019distilbert,q-bert,intel8b}. Generally, these methods require fine-tuning and often sacrifice accuracy. 
Quantization methods prolong training time by as much as $10\times$. 
It is for this reason that it is best to find model compaction methods that can work directly off the fine-tuned models.
}

We present a model compaction method, \OURL for such attention-based models. \OURL accepts as input a \textit{trained} model and reduces the number of bits needed to represent its parameters be it weights or embeddings. \OURL maintains accuracy without any further model refinement such as retraining or fine-tuning. Fine-tuning requires access to the dataset which may not be available or may be not possible under strict time constraints (daily updates to the model).


\OURL compacts the original floating-point parameters representing the vast majority of them with 3b or 4b. \OURL is plug-in compatible with any execution engine for transformer models as after decoding it, \OURL produces a model with identical architecture (layer dimensions, types, and connectivity) containing floating-point parameters. \OURL can be used as an off- and potentially on-chip compression method to reduce footprint, traffic and energy and thus amplify bandwidth, capacity, performance and energy efficiency. \OURL can further boost performance and energy efficiency as it can simplify computation converting most multiplications in additions. Accordingly, as a second use of \OURL, we present a specialized functional unit that takes advantage of \OURL's representation to reduce computation costs improving energy efficiency and performance.

We find that per layer the vast majority of weights closely follow some Gaussian distribution (whose parameters vary across layers), with very few --- typically less 0.1\% per layer --- ``outlier'' weights being the exception. 
\OURL capitalizes on this phenomenon using a twofold approach: First, it stores the few outlier weights as-is. 
Second, \OURL uses a dictionary of very few --- typically 8 --- representative values (centroids) for all other weights. As a result, it stores the vast majority of the weights --- typically 99.9\% --- as 3b indexes. \OURL uses a novel representative value selection algorithm  that results in higher accuracy and is much faster to converge compared to linear partitioning or K-Means.

Deep Compression~\cite{han2015deep} is representative of a class of dictionary-based compression methods that have been used to compress the  parameters of fixed-point models. It was originally demonstrated on 16b fixed-point convolutional neural networks. EIE takes advantage of Deep Compression to greatly boost energy efficiency when executing the resulting \textit{sparse} neural networks~\cite{EIE}. It stores weights as Huffman-encoded indexes into a dictionary of few representative values. The weights are expanded into their 16b fixed-point representative values before the MAC units. 
Compared to Deep Compression \OURL does not require fine-tuning, targets attention-based models which use floating-point values and where preserving outliers proves essential for maintaining accuracy, and uses a novel method for selecting the representative values. {In addition, the \OURL hardware architecture never expands the quantized weights to their representative floating-point values.} 

Outlier-aware quantization targets fixed-point convolutional models~\cite{outlierAware,outlierAwareHW}. Compared to \OURL, it  requires determining in advance what fraction of values should be outliers, uses a much larger fraction of outliers, typically, 3\%--5\%, uses linear quantization to reduce datawidth for the non-outliers, and requires fine-tuning of the model. Fang~\textit{et al.}, proposed a post-training quantization for convolutional neural networks~\cite{samsung20}. They observed that the values tend to follow a bell-shaped distribution which they take advantage using a piece-wise linear quantization. 
\OURL targets attention-based models which are floating-point based, automatically adjusts the fraction of outliers using a Gaussian distribution fit, utilizes a novel fast converging method for selecting the non-uniform quantization values for the non-outliers, and requires no fine-tuning.

We evaluate \OURL on various datasets and state-of-the-art attention-based NLP models: BERT (two variants), DistilBERT~\cite{sanh2019distilbert}, RoBERTa (two variants)~\cite{robetra}, HAT~\cite{HAT}, and SpanBERT~\cite{m2019spanbert}. We also compare \OURL with two quantized BERT models, Q8BERT~\cite{intel8b} and Q-BERT~\cite{q-bert}.


\null

 We highlight the following experimental findings:
 \begin{itemize} 
  \item For the most challenging task in the GLUE~\cite{glue}, MNLI, \OURL maintains accuracy while quantizing 99.9\% of the weights to 3 bits.
  \item Our centroid selection algorithm converges $9\times$ faster than K-Means selection and consistently reduces the number of required centroids to half.
  \item A practical implementation of \OURL compression for off-chip memory reduces model footprint by $10\times$. For TPU this translates to $10\times$ performance.
  \item Under iso-compute-area constraints, the \OURL accelerator is on average $7\times$ faster and consumes $3\times$ less energy than an accelerator with TensorCore-like units.
\end{itemize}
\null
\section{The BERT Family of NLP Models}
\label{sec:models}
\sloppy{
Google's BERT~\cite{BERT} (Bidirectional Encoder Representations from Transformers) is an attention-based model that is the model of choice for a variety of NLP tasks.}

\noindent\textbf{BERT-Large and BERT-Base: }Training deep learning models capable of  state-of-the-art accuracy on NLP tasks is expensive in terms of time, energy and volume of data.  \textsc{nvidia} reports that to train BERT from scratch in less than 1 hour, a cluster with 1,472, V100 GPUs was required~\cite{narasimhan_2019}. Each V100 GPU has 32GB of memory and consumes 450W at a total power cost of $662KW$ for the GPUs alone. To enable practical deployment for several tasks without access to such computing power and budgets the BERT framework introduced a ``pre-training and fine-tuning'' approach. BERT is pre-trained \textit{once} on an unlabeled dataset ---billions of words--- and then the pre-trained model can be fine-tuned for a few epochs to perform various tasks. Two pre-trained BERT models have been released: \textit{BERT-Base} and \textit{BERT-Large}. BERT-Large typically achieves higher accuracy leveraging $3.5\times$ more parameters. 

\noindent\textbf{Tasks: }BERT is most useful in language understanding tasks, such as sentiment analysis, paraphrasing, sentence similarity detection, and question answering. GLUE~\cite{glue} (General Language Understanding Evaluation) and SQuAD~\cite{squad} (Stanford Question Answering Dataset) are two popular benchmarks for such tasks. In this study, we focus mostly on the MNLI (The Multi-Genre Natural Language Inference) task of GLUE, since a)~it is the most comprehensive inference task in the dataset, and b)~it is the most sensitive task to quantization~\cite{BERTsurv}. MNLI given two provided sentences, the premise and the hypothesis, predicts if the premise entails the hypothesis, contradicts with it, or neither. As a representative of other less sensitive to quantization tasks in GLUE, we evaluate STS-B (Semantic Textual Similarity Benchmark) which tries to predict the similarity score between two human-annotated sentences. We use MNLI for all models since we are able to fine-tune BERT on the hardware that is available to us. Unfortunately, this is not possible for SQuAD. Its dataset is among the largest ones that are available and fine-tuning BERT for it is not practical on a single GPU. However, we had access to fine-tuned BERT-Large and SpanBERT models on SQuAD, and since our method does not require fine-tuning, we do evaluate \OURL on SQuAD for these models. We further evaluate \OURL on an English to French translation task using a Transformer model (HAT).

\begin{figure*}[t!]
\centering
\captionsetup{font=footnotesize,labelfont=footnotesize}
\subfloat[BERT layer architecture. Boxes are FC layers.]{\label{fig:BERT_ARCH}
\includegraphics[width=0.32\textwidth]{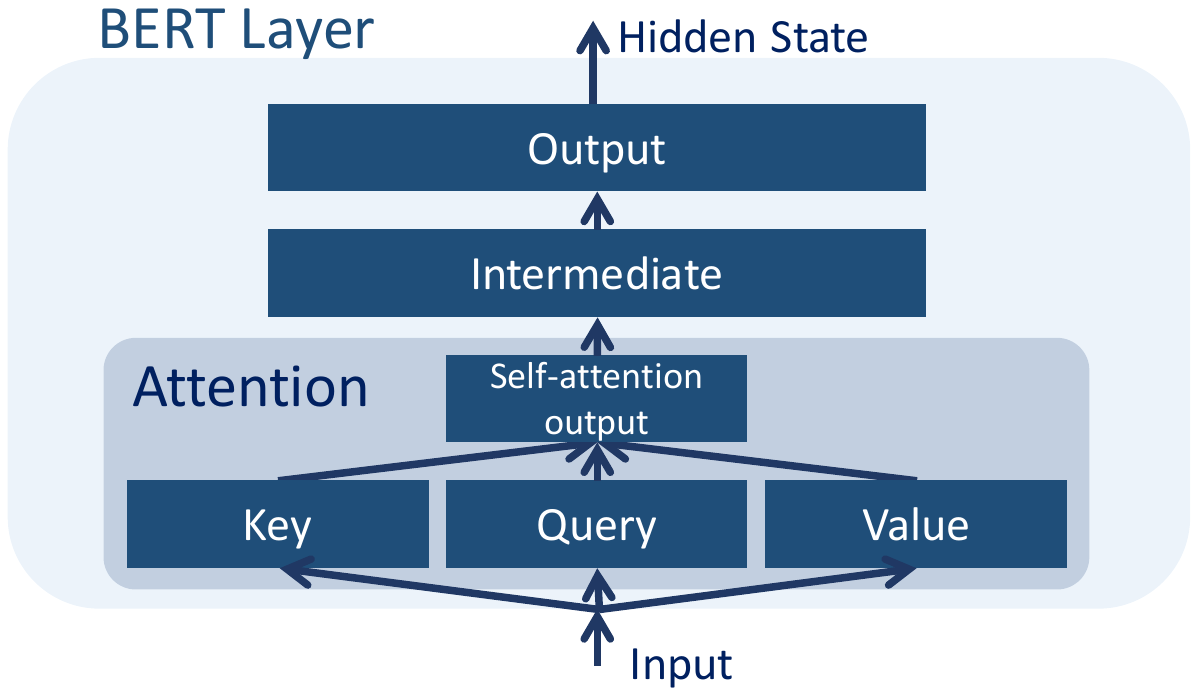}}
\subfloat[Per layer weight distribution]{\label{fig:BERT_Dist_out:a}\includegraphics[width=0.32\textwidth]{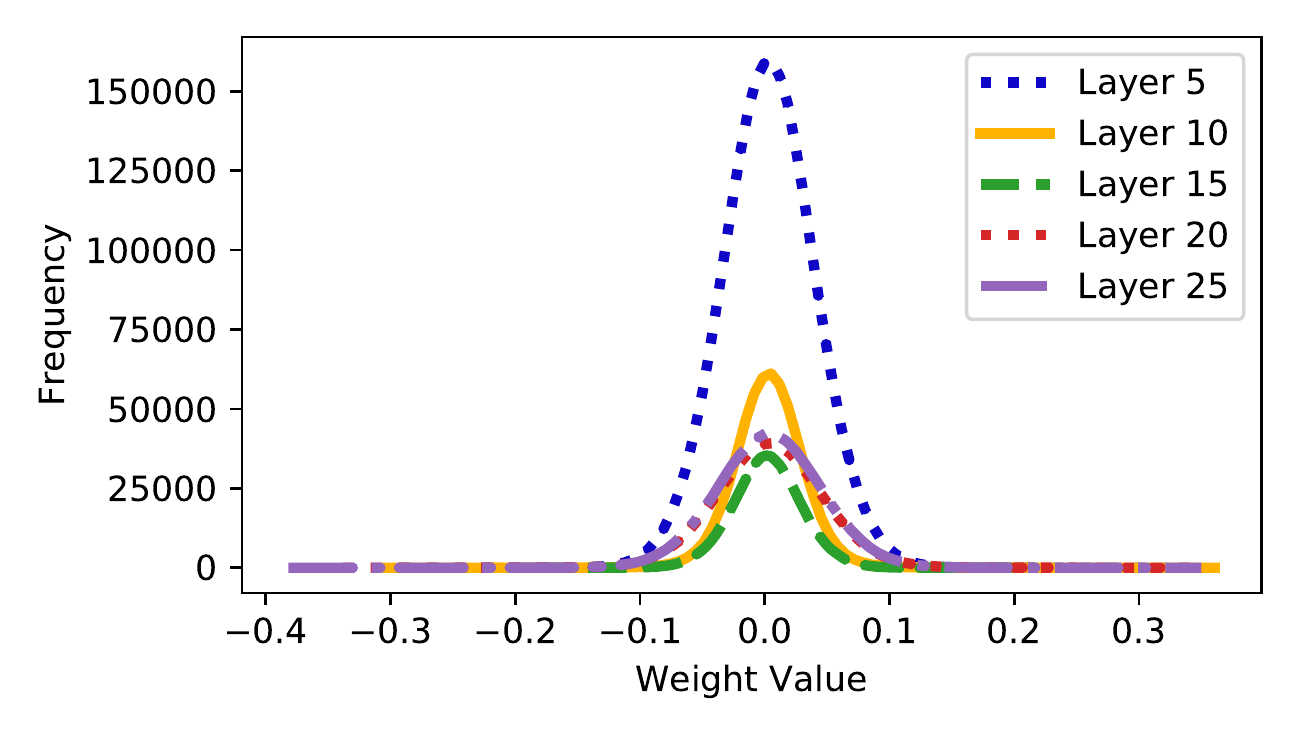} }%
\subfloat[Weights of a BERT layer]{\label{fig:BERT_Dist_out:b}
\includegraphics[width=0.32\textwidth]{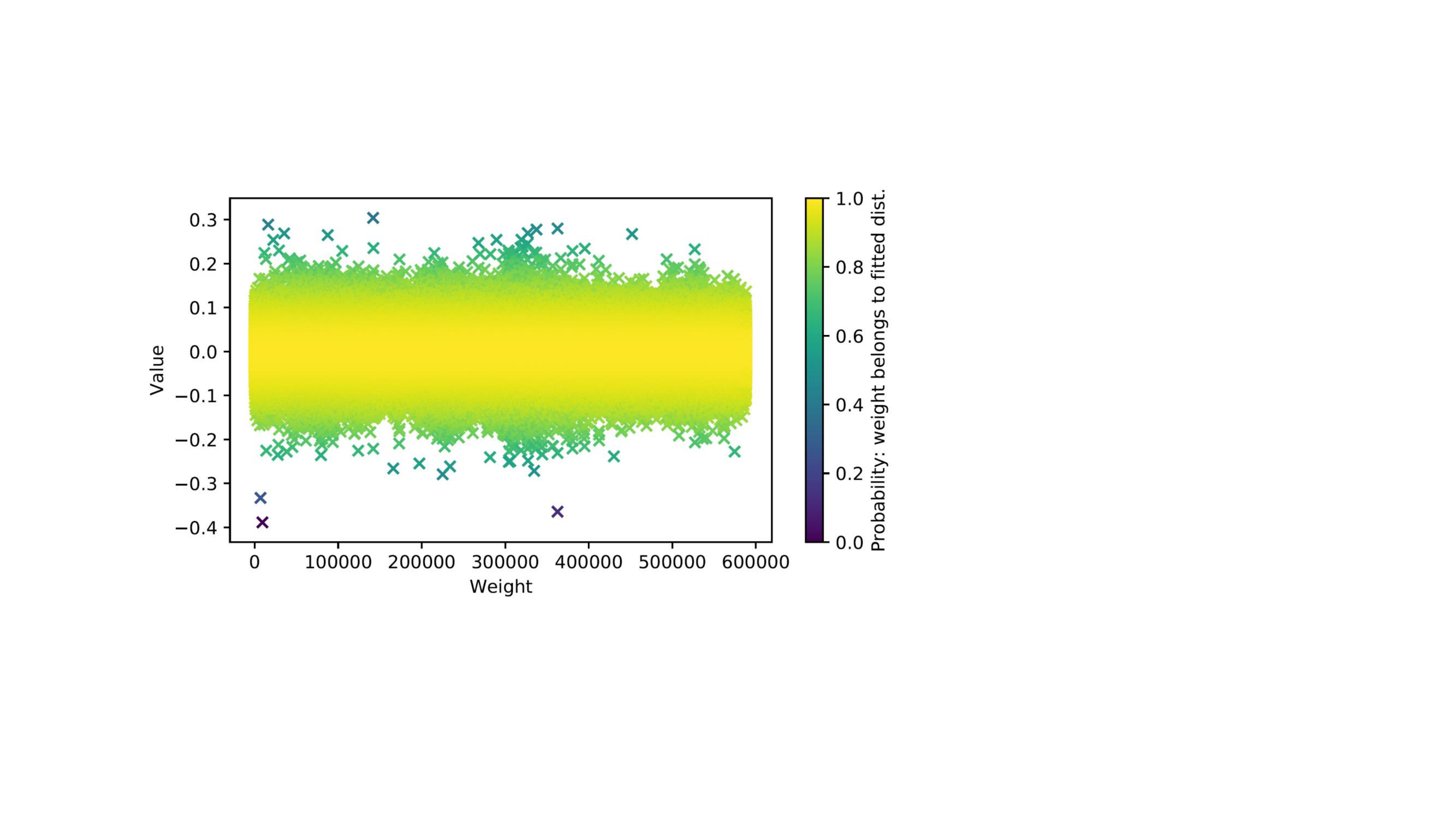} }%
\captionsetup{font=normalsize,labelfont=normalsize}
\caption{BERT layer architecture and per layer weight values.}
\end{figure*}

\noindent\textbf{BERT Architecture: }BERT-base consists of 12 BERT Layers while BERT-large has 24. Figure~\ref{fig:BERT_ARCH} shows that each BERT layer has 3 components: Attention, Intermediate, and Output. Each component includes a series of fully connected layers (FC) followed by a single normalization layer. The hidden state is a 768- to 1K-element vector (size varies per model and layer). After the last BERT layer, there is a single FC layer, the \textit{Pooler}. Table~\ref{tbl:BERT} details the configuration of these models. In total, BERT-Base has 73 ($\mathit{12 \times 6 + 1}$) FC layers and 110M parameters, whereas BERT-Large has 145 ($\mathit{24 \times 6 + 1}$) and 340M parameters. Activations and weights are 32b floating-point numbers and Table~\ref{tbl:BERT_Mem} reports the resulting memory footprint. Since BERT consists of mostly FC layers, and since the hidden state is a relatively short vector, it is the weights that dominate memory footprint and have to be streamed from off-chip. A set of embedding tables map the raw input into the vectors the network uses.

\begin{table}
\centering
\caption{BERT Architecture}
\label{tbl:BERT}
\scriptsize
\ra{1.2}
  \renewcommand{\familydefault}{\sfdefault}\normalfont

\begin{tabular}{ccc|cc}
                 & \multicolumn{2}{c}{\textbf{BERT-Base}} & \multicolumn{2}{c}{\textbf{BERT-Large}}  \\ \hline
BERT layers & \multicolumn{2}{c}{12}        & \multicolumn{2}{c}{24}          \\ \hline
Component        & FC \#       & Dimensions            & FC \# & \multicolumn{1}{c}{Dimensions} \\ \hline
Attention        & 4x          & 768 x 768       & 4x    & 1024 x 1024              \\
Intermediate     & 1x          & 768 x 3072      & 1x    & 1024 x 4096              \\
Output           & 1x          & 3072 x 768      & 1x    & 4096 x 1024              \\ \hline
BERT Pooler      & \multicolumn{2}{c}{768 x 768} & \multicolumn{2}{c}{1024 x 1024}
\end{tabular}

\hspace{2pt}
  \renewcommand{\familydefault}{\rmdefault}\normalfont

\scriptsize
\caption{BERT Memory Footprint}
\label{tbl:BERT_Mem}
\ra{1.2}
  \renewcommand{\familydefault}{\sfdefault}\normalfont

\begin{tabular}{c|cc}

Model               & \textbf{BERT-Base}   & \textbf{BERT-Large} \\ \hline
Embedding Tables    & 89.42 MB    & 119.22 MB   \\ \hline
Weights             & \textbf{326.26 MB}   & \textbf{1.12 GB}    \\ \hline
Model Input per Word      & 3 KB        & 4 KB       \\ \hline
Largest Layer Acts per Word & 12 KB       & 16 KB      \\
Sequence Length     & 128         & 128        \\
Activations         & 1.5 MB      & 2 MB       \\ \hline
\end{tabular}
\end{table}

\noindent\textbf{BERT Derivatives: }
BERT variants have been since released improving accuracy or reducing size. 
DistilBERT~\cite{sanh2019distilbert} uses knowledge distillation over the pre-trained BERT models to train a smaller, yet similar architecture. Facebook's RoBERTa~\cite{robetra} uses hyperparameter tuning, a different training method and a different embedding table to improve accuracy while maintaining the same architecture.  We also compare \OURL to two state-of-the-art quantized BERT variants, Intel's Q8BERT~\cite{intel8b} which uses 8b fixed-point values, and Q-BERT~\cite{q-bert} which uses dictionary compression. 

\noindent\textbf{Per Layer Weight Distribution: } 
Figure~\ref{fig:BERT_Dist_out:a} shows the distribution of the weights for a few layers of BERT-Base which are representative of the shape of the distributions for the other layers. In every layer the weights closely follow a Gaussian distribution which can be described by the mean and standard deviation of the layer's weights. The Gaussian-like distribution of parameters was observed in various DNNs~\cite{samsung20,G_DNN, G_DNN2, gordon2020compressing}.   Figure~\ref{fig:BERT_Dist_out:b} shows a color coded representation of a layer's weights (one per x coordinate) based on the probability of each weight belonging to the layer's Gaussian distribution. This reveals that there is a tiny fraction of weights that are on the fringes of the Gaussian distribution as their magnitude is considerably larger than the rest of the weights. 

These observations motivate \OURL which splits the weights of each layer into two groups. The ``G'' (Gaussian) group consists of weights whose magnitude fits within 99.9\% of the values as described by the Gaussian distribution formed by the mean and standard deviation of all weights in the layer. The second group, the ``Outliers'' (``O'') includes values that fall outside the Gaussian distribution. We have found experimentally that:
\begin{enumerate*} 
  \item representing just the outliers precisely and quantizing the rest of the model to a few representative values (e.g, 8) is sufficient for maintaining model accuracy.
  \item Using representative values for all weights either drastically reduced compression (too many representative values were needed) or sacrificed accuracy.
\end{enumerate*}

\section{Related Work}
\label{sec:related}

Compression methods for NLP models fall under three different approaches, Model Quantization, Pruning, and Knowledge Distillation. In model quantization the objective is to reduce the number of bits for representing the model's parameters while keeping the model architecture as-is. Pruning's goal is to remove some of the weights by forcing them to be zero. Combining pruning with zero-aware memory encoding often reduces the model's footprint. Knowledge distillation trains a smaller model, the student, to mimic the behaviour of a much larger teacher model. 

\noindent\textbf{Model Quantization: }Quantization techniques can be direct or indirect. Direct methods map the weights to a fixed-point representation, whereas indirect methods use a dictionary of representative values and encoded weights as indexes.

Intel's Q8BERT introduces a fine-tuning method to quantize the weights and activations to 8-bit fixed-point values~\cite{intel8b}. Some operations such as Softmax and Layer Normalization are not quantized and use FP32. We compare the accuracy of \OURL with Q8BERT on the MNLI task in Section~\ref{SEC:RESULT_BERT_BASE} showing that \OURL reduces model size more than Q8BERT while maintaining accuracy. 
Furthermore, \OURL is faster to deploy since it does not require fine-tuning. 
Although, the decompressed model with \OURL uses FP32 values, the \OURL hardware accelerator performs most computations without decompressing the weights. 


Q-BERT is a dictionary based fine-tuning approach that uses second order Hessian information to quantize the model's weights to a few (4 to 16) representative values. It stores weights as indexes to those values~\cite{q-bert}. Q-BERT separates the weights of each layer into multiple groups and quantizes each group separately using per group dictionaries each of 4, 8 or 16 entries. Dividing each layer in 128 groups results in acceptable accuracy. Finally, Q-BERT quantizes the embedding tables to 8b to avoid a significant loss in accuracy. \OURL does not require fine-tuning, only keeps one dictionary per layer and quantizes the embedding layers to 3b. Section~\ref{SEC:RESULT_BERT_BASE} shows that \OURL achieves higher compression than Q-BERT while maintaining accuracy.

\noindent\textbf{Model Pruning: }Weight pruning reduces model footprint by forcing a portion of the weights to be zero. 
Gordon~\textit{et al.}, studied pruning BERT during training~\cite{gordon2020compressing}. They showed that 30\%-40\% of the weights based on the magnitude could be pruned with minimal effect on the accuracy of the final task. They found that MNLI was the task that was most sensitive to pruning.
Structured pruning removes a series of weights that correspond to a component of the model~\cite{BERTsurv}. Attention head pruning~\cite{darksecret,raganato2020fixed} and Encoder unit pruning~\cite{fan2019reducing} are examples of this approach. Pruning methods require fine-tuning to compensate for the initial accuracy loss. As we will show, \OURL achieves nearly $10\times$ compression (99.9\% of 32b values are compressed to 3b). Even if we ignore its encoding overhead, a pruning method should remove nearly 90\% of the weights to achieve similar compression. \OURL could complement pruning an investigation left for future work.

\noindent\textbf{Knowledge Distillation: }Knowledge distillation trains a smaller model (student) from a larger model (teacher). Based on what the student learns from the teacher there are three groups of Knowledge distillation approaches for BERT~\cite{BERTsurv} currently in use. In the first group, the student learns the behaviour of the encoder layer. The student can have fewer attention heads in each layer~\cite{zhao2019extreme,sunmobilebert,jiao2019tinybert} or fewer encoder layers~\cite{sanh2019distilbert,sun2019patient}. Another approach trains a student based on the output \textit{logits} (input of the last layer's softmax)~\cite{cheng2017survey}. Furthermore, The student is free to adopt a different type of network or components thereof such as a Convolutional Neural Network (CNN) or Long Short Term Memory (LSTM). Tang~\textit{et al.}, used a Bidirectional LSTM (BiLSTM) based architecture to replace the attention layers~\cite{tang2019distilling, tang2019natural}. Chia~\textit{et al.}, proposed a CNN-based model to replace transformer layers where the student tries to learn the contextual dependence of the tokens (words) from the output of each attention layer~\cite{chia2019transformer}. DistilBERT~\cite{sanh2019distilbert} is distilled model of BERT-Base and is about half in size. Section~\ref{SEC:EVAL_DISTILI} shows that \OURL compresses DistilBERT by $10\times$ and results in a model that is $20\times$ smaller than BERT-Base.

\section{\OURLCORE\ Quantization}\label{sec:method}

For BERT, \OURL operates at the granularity of a layer and over the fine-tuned model. The compaction process starts with separating the weights into two groups, the ``Gausian'' (G) and the ``Outliers'' (O). \OURL stores the outliers as-is (FP32) whereas it quantizes the ``G'' group values to a few representative values. Only a tiny fraction of the weights, typically less than 0.1\%, end up in the ``O'' group. \OURL reduces overall model size by quantizing the Gaussian group. Since the weight distribution is not uniform, we propose a non-linear quantization method that results in higher resolution where the weights are densely populated. This method proves capable of encoding roughly 99.9\% of the weights with just 8 representative FP32 values, while the inference error is kept below 1\% (or with 16 representative values for no accuracy loss). A ``G'' group weight is stored as a 3b index to those representative values. \OURL uses just \textit{one} set of representative values per layer.


In summary, \OURL stores the following information per layer:
\begin{enumerate*} 
  \item The outliers in original form. (FP32)
  \item The bin index for each ``G'' group weight. (3b)
  \item A reconstruction table for weights which represents the representative values (\textit{centroids}) for each bin. (FP32)
\end{enumerate*}
We defer the description of a practical encoding until Section~\ref{sec:memcompression} and after we discuss the ``G'' and ``O'' group selection and quantization process.

\subsection{Outlier Detection}
To detect the outlier weights for an FC layer, \OURL first computes the mean and the standard deviation of the layer's weights. Then per weight, it computes a probability that the weight belongs to that distribution using the \textsc{\textit{PDF}} (Eq.~\ref{EQ:pdf}) where $x$ is the weight, $\mu$ is the mean, and $\sigma$ is the standard deviation. \OURL uses a threshold, a configuration parameter, to select outliers. A weight whose probability is less than the threshold is deemed as an outlier. We empirically found that a log-probability threshold of \textit{{-}4} is sufficient for maintaining overall accuracy.
\begin{equation}\footnotesize\label{EQ:pdf}
\mathit{pdf}(x|\mu ,\sigma ^2)= \frac{1}{\sqrt{2\pi \sigma ^2} } e^{-\frac{(x-\mu)^2}{2\sigma^2}}
\end{equation}
The closest prior work to \OURL's outlier selection approach is outlier-aware quantization~\cite{outlierAware,outlierAwareHW}. This method targets fixed-point CNNs and considers a fixed, predetermined fraction of the weights (at least 3\% which is an order of magnitude more than \OURL) as outliers. The outliers remain in 16b and non-outlier values are linearly quantized to 4b. To reduce the quantization error, fine-tuning is required. With \OURL's nonlinear quantization having about 1 outlier every 1000 weights proves sufficient for preserving accuracy without fine-tuning. In addition, \OURL uses dictionary-based compression of the non-outlier group.
Section~\ref{Eval:Quant} compares \OURL with an outlier-aware inspired method that uses Linear Quantization for the non-outliers.



\subsection{``G'' Group Weight Quantization}
\OURL's aims to represent the ``G'' weights with few representative FP32 values. This amounts to clustering the weights and assigning a representative value per cluster. 
We use clusters of equal population. Intuitively, this objective will put more clusters where weights are densely populated affording higher resolution where there are more weights. This approach can be implemented by sorting the weights and dividing them to equally sized clusters. The first and the last weight in each cluster determine the boundaries of that cluster. Then the average of the weights inside each cluster can be used as the centroid. We find that quantizing BERT-Base with this approach into 8 clusters (3b indexes) degrades inference accuracy by 10\% in GLUE tasks. 

To reduce this error \OURL uses an iterative approach that bears similarity to K-Means. The process repeatedly applies the following two steps: 1)~\OURL~moves a weight from cluster A to cluster B if the $L1$ distance of the weight and the centroid of cluster A is greater than the $L1$ distance of the weight and the centroid of cluster B. 2)~After re-assigning the weights to clusters, \OURL updates the centroids by computing the new average over the weights of each cluster. This is equivalent to minimizing the $L2$ distance.
This iterative process repeats until the sum of $L1$ distances between centroids and weights is minimized. For a 3b quantization, we observed that 7 iterations are enough to converge to the optimal solution. 

\begin{figure*}[t!]

\centering

\begin{minipage}{0.34\textwidth} 
\includegraphics[width=.96\textwidth]{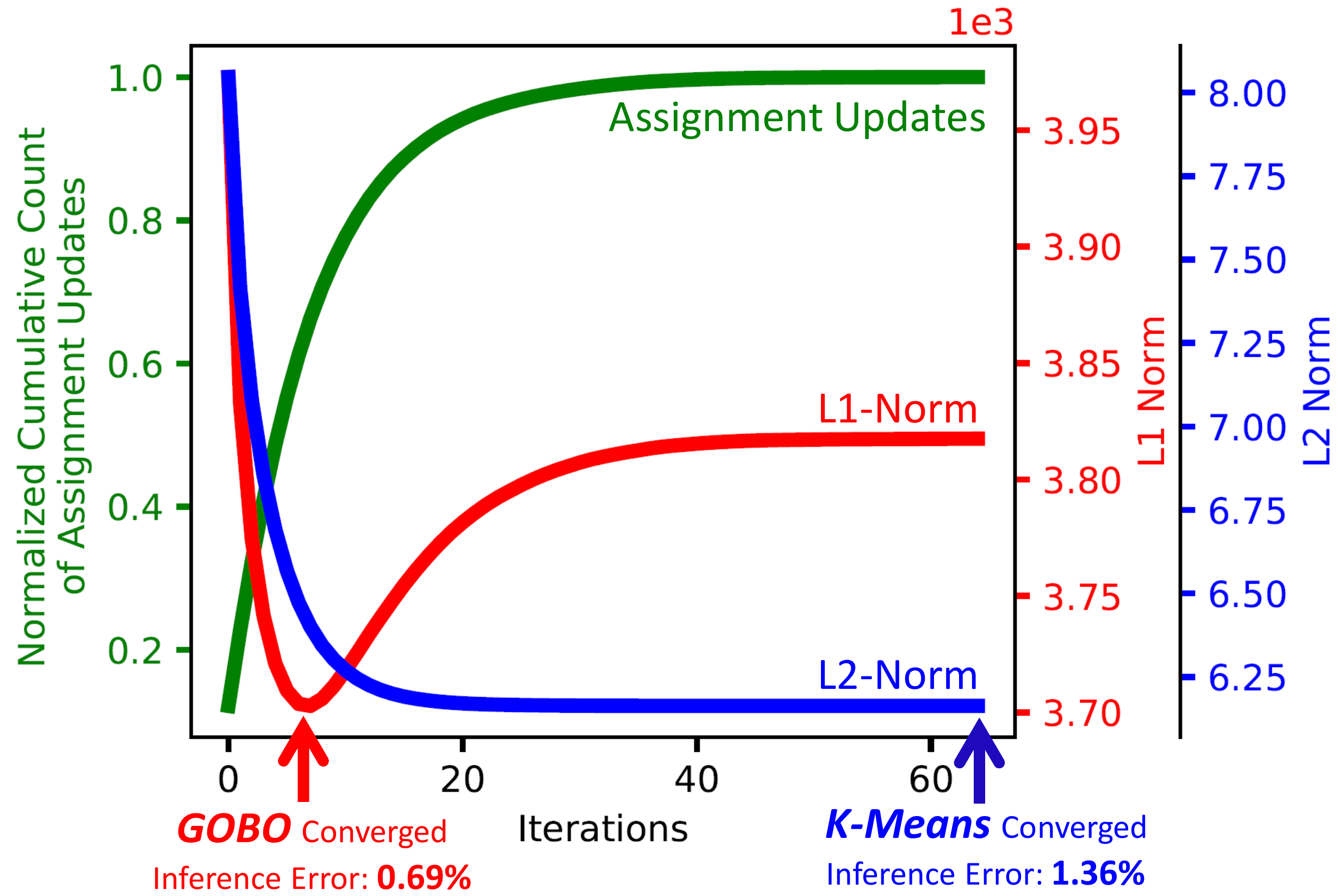}
\captionsetup{font=footnotesize,labelfont=footnotesize}
\caption{\OURL~and K-Means convergence.} 
\label{fig:L1L2}
\end{minipage}\hspace{0pt}
\begin{minipage}{0.31\textwidth}
\centering
\includegraphics[width=.85\textwidth]{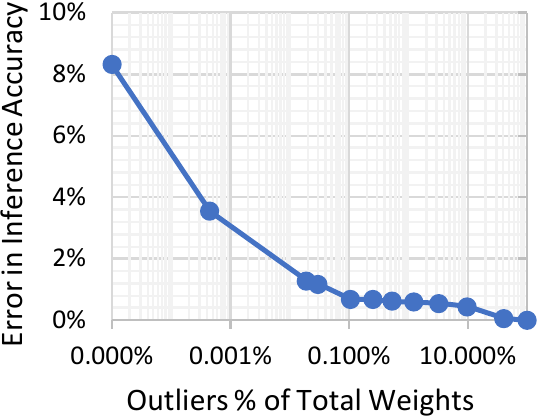}
\captionsetup{font=footnotesize,labelfont=footnotesize}
\caption{Outlier's impact on inference accuracy of BERT-Base (MNLI)}
\label{fig:outlier_impact}
\end{minipage}\hspace{5pt}
\begin{minipage}{0.31\textwidth} 
\centering
\includegraphics[width=.9\textwidth]{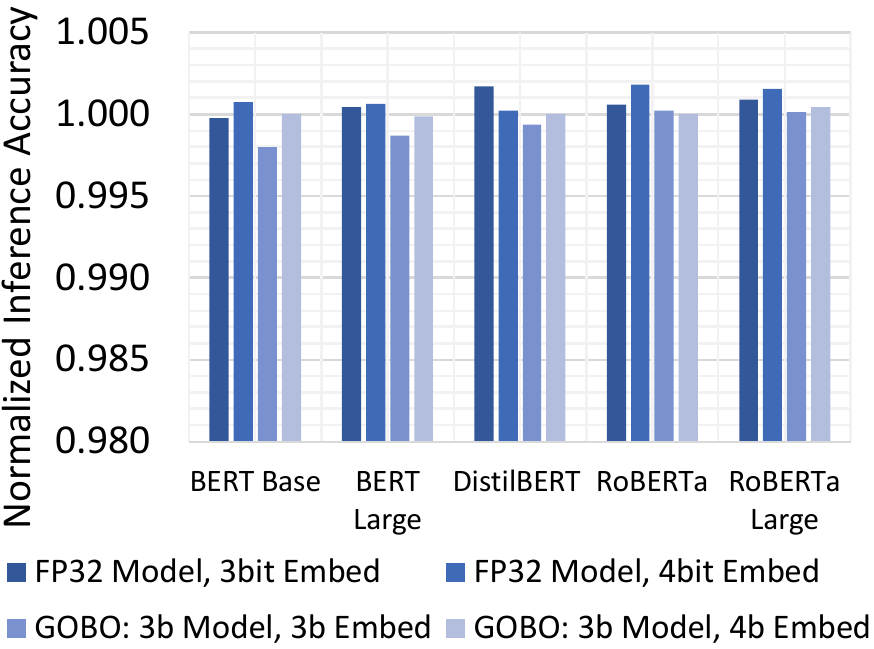}
\captionsetup{font=footnotesize,labelfont=footnotesize}
\caption{Effect of Embedding Table Quantization on Accuracy.}
\label{fig:EmbedQuant}
\end{minipage}
\end{figure*}

Terminating the process using the $L1$ proves faster and results in a model with higher inference accuracy using the $L2$.
Figure~\ref{fig:L1L2} helps explaining why this is so. The figure shows how the $L1$ and $L2$ evolve over time (iterations) and also the cumulative distribution of weight-to-bin reassignments. 
Weight reassignments follow Zipf's law with more than 80\% of them happening in the first 15\% of iterations. This explains the initial rapid drop in $L1$ which the first step of each iteration directly minimizes. Since reducing the $L1$ also reduces $L2$ (but not vice versa) this also improves $L2$. The second step reassigns centroids to minimize $L2$ instead. As the figure shows, this reassignment proceeds much more slowly and quickly hurts $L1$ as it overemphasizes weights that are further away from the centroid. Given the Gaussian distribution, these weights disproportionately affect centroid selection reducing representation accuracy for all weights. Terminating when $L1$ is minimized better balances the per value representation accuracy. Empirically, emphasizing all weights equally rather the overemphasizing a few, results in better overall inference accuracy. \OURL's approach converges $9\times$ faster than K-Means and as Section~\ref{Eval:Quant} describes achieves higher accuracy.

Deep Compression uses dictionary compression for CNNs utilizing K-Means with linear initialization for cluster centroids and requiring fine-tuning to regain any accuracy loss~\cite{han2015deep}. It minimizes the $L2-Norm$ inside each cluster. \OURL maintains the model's accuracy without any help from retraining. The centroid initialization in \OURL is nonlinear and depends on the per layer weight distribution. \OURL detects a few but effective outliers and keeps them in their original representation. importantly, \OURL objective is to minimize $\LNorm{1}$ within each cluster rather than $L2$. Algorithm~\ref{Algo} summarizes how \OURL compacts each layer.





\begin{algorithm}
\footnotesize
 Compute the mean$(m)$ and standard deviation$(sd)$ of the weights\;
 Detect outliers based on the Gaussian dist. specified by $m$ and $sd$\;
 Store outlier weights as-is\;
 Sort the ``G''  weights and place them into bins with equal capacity\;
 Compute the min, max, and mean (centroid) of each bin\;
 $L1_{new} \gets$ L1-Norm of the current assignment\;
 \Do{$L1_{old} \geq L1_{new}$}{
    $L1_{old} \gets L1_{new}$\;
    Reassign weights the closest centroid\;
    Update centroids based on the new assignment\;
    $L1_{new} \gets$ L1-Norm of the current assignment\;
     }
 \caption{\OURL Quantization}
 \label{Algo}
\end{algorithm}

\subsection{Accuracy and Compression Potential}
\label{Eval:Quant}

This section evaluates \OURL's effect on accuracy and compression rate. Here we ignore the additional information that any practical encoding will require. We introduce a specific encoding in Section~\ref{sec:memcompression} where we use \OURL for off-chip memory compression. As it will be shown, the overhead is tiny which is no surprise as it includes the representative values and the coordinates of the outliers.

\noindent\textbf{Methodology: }We use the pre-trained models, the scripts and datasets provided by the Hugging Face repository~\cite{Wolf2019HuggingFacesTS} and the SciKit-Learn library. Specifically, to fit a Gaussian distribution we use \textit{sklearn.GaussianMixture} with one Gaussian component. Then the log probability for each weight is computed using \textit{score-samples}. We considered the weights with the log probabilities of $\mhyphen 4$ or less as outliers. 

To show the advantage of \OURL's clustering approach we compare the accuracy of each model when the ``G'' weights are quantized by Linear Quantization and K-Means. In linear quantization, the range of non-outlier weights (``G'' group) is linearly divided by the number or bins ($2^{Bits}$). In K-Means we use the same centroid initialization as \OURL and iterate until the cluster assignments converge. The outlier weights in all of these methods are detected and represented in the same manner.
All experiments we performed on a dual RTX~2080~Ti 11GB GPU system with an AMD Ryzen Threadripper 1920X CPU with 128GB Quad-channel DDR4-3200 RAM.

\noindent\textbf{{Importance of Outliers:} }
\label{sec:gobo:outliers}
Figure~\ref{fig:outlier_impact} reports the absolute loss in inference accuracy of 3-bit quantized BERT-Base on MNLI when the fraction of outliers is swept from 0\% to 100\%. \OURL~without outliers incurs an 8.3\% accuracy loss. Having just 0.1\% of outliers reduces this loss to less than 1\%. Increasing the number of outliers by $100\times$ to 10\% improves accuracy by just 0.25\%. 


\noindent\textbf{BERT-Base: }
\label{SEC:RESULT_BERT_BASE}
First we compare \OURL to other BERT-specific quantization methods. In this section we focus on the MNLI task which is the most challenging among the GLUE set of tasks.
Table~\ref{tbl:Relwork} shows accuracy on the MNLI task with different quantization methods: Intel's Q8BERT~\cite{intel8b}, Q-BERT~\cite{q-bert}, and \OURL with 3b or 4b ``G'' weights. Recall that Q-BERT and Q8BERT require fine-tuning. Using the Hugging Face PyTorch library, fine-tuning BERT-Base for the MNLI task on our dual GPU system took about 3 hours per epoch. We fine-tuned the model for 3 epochs as suggested by the Hugging Face documentation. After 3 epochs (9 hours) we achieved the baseline accuracy of 84.45\%. Fine-tuning the same task on the same machine for Q8BERT, takes about 34 hours per epoch. Based on the authors' suggestion, we trained the model for 3 epochs (102 hours). Quantizing the same model with \OURL takes about 10 minutes using a single CPU core of our system. 

The results show that \OURL can compact the model by $9.8\times$ and with less than 0.7\% loss in accuracy and by $7.92\times$ with no accuracy loss. Q-BERT compacts the model by $6.5\times$ at a $0.56\%$ accuracy loss or by $7.81\times$ with a $1.04\%$ accuracy loss. Q8BERT reduces the model the least and by $4\times$ at a $0.7\%$ accuracy loss. Recall, however, that it uses 8b fixed-point. In summary, \OURL produces models that are smaller and with similar or no accuracy loss, and produces the model within minutes vs. days.

We next show that our proposed centroid selection method for the ``G'' group is better than two prior methods: linear quantization and K-Means. We study \OURL with our proposed quantization, \OURL with linear quantization, and \OURL with K-Means quantization. We study how accuracy is affected as we change the number of centroids (all experiments have the same set of outliers). Tables~\ref{tbl:MNLI} shows that \OURL as proposed when using 3b weights (8 centroids) incurs an accuracy loss of 0.69\% which is considerably less than then 1.36\% loss incurred with K-Means. Linear quantization performs the worst incurring an error of nearly 52\%.  To maintain the baseline accuracy, \OURL, K-Means and Linear Quantization require, 4b, 5b, and 6b weight indexes respectively.  We note that using 4b weight indexes amounts to a 33\% increase over using 3b. Moreover, our proposed selection method is faster than K-Means.

Similar behavior is observed for the STS-B task as Table~\ref{tbl:STSb} shows. As expected STS-B is less sensitive to quantization. \OURL incurs no accuracy loss with 3b, whereas K-Means needs 4b and linear quantization requires 5b.

\begin{table*}[t!]
\centering
\scriptsize
\caption{Comparison of \OURL and BERT-Specific Quantization Methods. BERT-Base on MNLI.}
\label{tbl:Relwork}
\ra{1.15}
  \renewcommand{\familydefault}{\sfdefault}\normalfont
\begin{tabular}{c|cccccc}
\textbf{BERT-Base (MNLI)} & \textbf{Weights}    & \textbf{Embedding } & \textbf{Accuracy (m)}     & \textbf{Error}           & \textbf{No Fine-tuning} & \textbf{Compression Ratio }    \\ \toprule
\textbf{Baseline}         & FP32       & FP32       & 84.45\%          & -               & -                       & 1$\times$             \\ \hline
Q8BERT~\cite{intel8b}           & 8-bit          & 8-bit          & 83.75\%          & 0.70\%          & \xmark                   & 4$\times$             \\
Q-BERT~\cite{q-bert}           & 3-bit         & 8-bit          & 83.41\%          & 1.04\%          & \xmark                   & 7.81$\times$          \\
Q-BERT~\cite{q-bert}           & 4-bit          & 8-bit          & 83.89\%          & 0.56\%          & \xmark                   & 6.52$\times$          \\
\OURL             & \textbf{3-bit} & \textbf{4-bit} & \textbf{83.76\%} & \textbf{0.69\%} & \cmark                   & $\textbf{9.83}\mathbf{\times}$ \\
\OURL             & \textbf{4-bit} & \textbf{4-bit} & \textbf{84.45\%} & \textbf{0.00\%} & \cmark                   & $\textbf{7.92}\mathbf{\times}$
\end{tabular}
\vspace{8pt}
\ra{1.1}
  \renewcommand{\familydefault}{\rmdefault}\normalfont
\caption{\OURL w/ ``G'' Group Centroid Selection Policies: BERT-Base/-Large. }
\label{tbl:MNLI}
\label{tbl:STSb}
\label{tbl:squad}
\ra{1.15}
  \renewcommand{\familydefault}{\sfdefault}\normalfont\begin{tabular}{c|c|cc|cc|cc|c}
                          & \multirow{1}{*}{\textbf{Baseline}} & \multicolumn{2}{c|}{\textbf{\OURLCORE~w/ Linear Quant.}} & \multicolumn{2}{c|}{\textbf{\OURLCORE~w/ K-Means}} & \multicolumn{2}{c|}{\textbf{\OURLCORE}} & \multirow{1}{*}{\textbf{Potential Comp. Ratio}} \\
                               \hline \multicolumn{9}{c}{\textbf{ GLUE/MNLI with BERT-Base}}\\\hline
\multicolumn{1}{c|}{Bits} &                           & Accuracy (m)          & Error            & Accuracy (m)          & Error            & Accuracy (m)   & Error    & Potential Comp. Ratio                                            \\ \hline
\multicolumn{1}{c|}{32}   & 84.45\%                   & -                     & -                & -                     & -                & -              & -        & 1$\times$                                            \\ \hline
\multicolumn{1}{c|}{2}    & -                         & 31.81\%               & 52.64\%          & 69.98\%               & 14.47\%          & 71.02\%        & 13.43\%  & 16$\times$                                   \\
\multicolumn{1}{c|}{3}    & -                         & 32.48\%               & 51.97\%          & 83.09\%               & 1.36\%           & 83.76\%        & \textbf{0.69\%}   & 10.67$\times$                                \\
\multicolumn{1}{c|}{4}    & -                         & 82.75\%               & 1.70\%           & 84.01\%               & \textbf{0.44\%}           & 84.45\%        & 0.00\%   & 8$\times$                                    \\
\multicolumn{1}{c|}{5}    & -                         & 84.20\%                & \textbf{0.25\%}           & 84.45\%               & 0.00\%           &                &          & 6.4$\times$                                  \\
\multicolumn{1}{c|}{6}    & -                         & 84.45\%               & 0.00\%           &                       &                  &                &          & 5.3$\times$\\                                
     \hline \multicolumn{9}{c}{\textbf{ GLUE/STS-B on BERT-Base}}\\\hline
Bits &                           & Spearman               & Error               & Spearman       & Error      & Spearman      & Error     &      Potential Comp. Ratio                                     \\ \hline
32   & 88.33\%                   & -                      & -                   & -              & -          & -             & -         & 1$\times   $                                         \\ \hline
2    & -                         & 2.67\%                 & 85.66\%             & 81.12\%          & 7.21\%       & 82.66\%         & 5.67\%      & 16$\times$                                              \\
3    & -                         & 74.00\%                   & 14.33\%             & 88.11\%          & 0.22\%       & 88.33\%         & \textbf{0.00\%}      & 10.67$\times  $                                         \\
4    & -                         & 87.46\%                & 0.87\%              & 88.33\%          & \textbf{0.00\%}       &               &           & 8$\times$                                               \\
5    & -                         & 88.33\%                &\textbf{ 0.00\% }             &                &            &               &           & 6.4$\times$\\                                            

          \hline \multicolumn{9}{c}{\textbf{SQuAD v1.1 with BERT-Large}}\\\hline
Bits &                           & F1 Score            & Error              & F1 Score      & Error       & F1 Score     & Error      &          Potential Comp. Ratio                                     \\ \hline
32   & 91.95\%                   & -                   & -                  & -             & -           & -            & -          & 1$\times$                                    \\ \hline
2    & -                         & 0.01\%              & 91.94\%            & 34.83\%       & 57.12\%     & 56.22\%      & 35.73\%    & 16$\times$                                   \\
3    & -                         & 5.37\%              & 86.58\%            & 90.56\%       & 1.39\%      & 91.04\%      & \textbf{0.91\% }    & 10.67$\times$                                \\
4    & -                         & 89.11\%             & 2.84\%             & 91.24\%       &\textbf{ 0.71\% }     & 91.95\%      & 0.00\%        & 8$\times$                                    \\
5    & -                         & 90.61\%             & 1.34\%             & 91.92\%         & 0.03\%       &              &            & 6.4$\times$                                  \\
6    & -                         & 91.88\%             & \textbf{0.07\% }            & 91.95\%         &0.00\%             &              &            & 5.3$\times$                                  \\
7    & -                         & 91.95\%             & 0.00\%                &               &             &              &            & 4.57$\times$                                
\end{tabular}
\vspace{8pt}
\ra{1.1}
  \renewcommand{\familydefault}{\rmdefault}\normalfont
\caption{\OURL w/ ``G'' Group Centroid Selection Policies:  GLUE/MNLI on DistilBERT.}
\label{tbl:distilibert}
\ra{1.15}
  \renewcommand{\familydefault}{\sfdefault}\normalfont\begin{tabular}{c|c|cc|cc|c}
\multirow{2}{*}{\textbf{Bits}} & \textbf{Baseline}     & \multicolumn{2}{c|}{\textbf{\OURLCORE~w/ K-Means}} & \multicolumn{2}{c|}{\textbf{\OURLCORE}} & \multirow{1}{*}{\textbf{Potential} } \\
                      & \textbf{Accuracy (m)} & \textbf{Accuracy (m)}     & \textbf{Error}     & \textbf{Accuracy (m)}   & \textbf{Error}   &    \textbf{Comp. Ratio}                                          \\ \toprule
32                    & 81.98\%      & -                & -         & -              & -       & 1$\times$                                    \\ \hline
3                     & -            & 80.83\%          & 1.15\%    & 81.30\%        & \textbf{0.68\%}  & 10.67$\times$                                \\
4                     & -            & 81.78\%          & \textbf{0.20\%}    & 81.98\%        & 0.00\%  & 8$\times$                                    \\
5                     & -            & 81.98\%          & 0.00\%    &                &         & 6.4$\times$                                 
\end{tabular}
\vspace{8pt}
\ra{1.1}
  \renewcommand{\familydefault}{\rmdefault}\normalfont
\caption{\OURL w/ ``G'' Group Centroid Selection Policies:  GLUE/MNLI on RoBERTa.}
\label{tbl:Roberta_base}
\label{tbl:Roberta_large}
\ra{1.12}
  \renewcommand{\familydefault}{\sfdefault}\normalfont\begin{tabular}{c|c|cc|cc|c|c|cc|cc|c}
\hline
\multicolumn{7}{c}{\textbf{RoBERTa-Base}} & \multicolumn{6}{|c}{\textbf{RoBERTa-Large}} \\
\hline
\multirow{2}{*}{Bits}                 & Baseline           & \multicolumn{2}{c|}{\OURL w/ K-Means} & \multicolumn{2}{c|}{\OURL} & \multirow{1}{*}{Potential}  & Baseline           & \multicolumn{2}{c|}{\OURL w/ K-Means} & \multicolumn{2}{c|}{\OURL} & \multirow{1}{*}{Potential} \\
                                      & Acc. (m)       & Acc. (m)    & Error      & Acc. (m)   & Error   &         CR      &   Acc. (m)       & Acc. (m)    & Error      & Acc. (m)   & Error   &                        CR                  \\ \toprule
32                                    & 87.60\%            & -               & -          & -              & -       & 1$\times$       & 90.20\%      & -                & -         & -              & -       & $1\times$                               \\   \hline
3                                     & \multirow{2}{*}{-} & 76.00\%         & 11.60\%    & 79.68\%        & 7.92\%  & 10.67$\times$   & -            & 82.18\%          & 8.02\%    & 84.26\%        & 5.94\%  & 10.67$\times$                               \\
3b/4b &                    &                 &            & 86.19\%        & \textbf{1.41\%}  & 10.13$\times$     & -            &                  &           & 89.33\%        & \textbf{0.87\%}  & 10.03$\times$                               \\
4                                     & -                  & 86.80\%         & 0.80\%     & 87.30\%        & 0.30\%  & 8$\times$          & -            & 89.48\%          & 0.72\%    & 89.88\%        & 0.32\%  & 8$\times$                                      \\
5                                     & -                  & 87.32\%         & 0.28\%     & 87.60\%        & 0.00\%  & 6.4$\times$         & -            & 90.07\%          & 0.13\%    & 90.20\%        & 0.00\%  & 6.4$\times$                                  \\
6                                     & -                  & 87.60\%          & 0.00\%     &                &         & 5.3$\times$      & -            & 90.20\%          & 0.00\%    &                &         & 5.3$\times$    \\
\end{tabular}
\end{table*}

\noindent\textbf{BERT-Large: }To our knowledge, ours is the first attempt to quantize BERT-Large on the SQuAD task. SQuAD is a complex task that requires days of fine-tuning when implemented over BERT-Large. We apply \OURL after this fine-tuning phase at a negligible cost in time. 
Table~\ref{tbl:squad} reports the compression and accuracy of the model with different ``G'' group quantization policies. Our centroid selection policy proves best. With 3b weight indexes the accuracy loss is less than 1\% and with 4b there is none.
For MNLI, \OURL incurs no accuracy loss even with 3b weight indexes. In the rest of this section we limit attention only to \OURL w/ K-means and \OURL with our proposed centroid selection policy.

\noindent\textbf{DistilBERT:}
\label{SEC:EVAL_DISTILI}
Table~\ref{tbl:distilibert} shows accuracy when quantizing DistilBERT which was distilled from BERT-Base and is about $2\times$ smaller. \OURL incurs no or less than 1\% accuracy loss with 4b and 3b weights respectively resulting in models that are $16\times$ or $21\times$ smaller than BERT-Base. In either case, K-Means requires twice as many bins.

\noindent\textbf{RoBERTa:} Table~\ref{tbl:Roberta_base} shows accuracy and compression ratio. Quantizing to 3b ``G'' weights incurs an accuracy loss of 8\%. Two FC layers (\textit{"Value layer"} in self-attention and \textit{Intermediate layer}) in the first 6 BERT Encoders are the most sensitive to quantization. We can either quantize the whole model to 4b ``G'' weights (accuracy loss of just 0.6\%), or better, we can use 4b ``G'' weights just for these two layers for the first 6 out of the 12 total Encoder layers and 3b for the rest. This reduces the accuracy loss to just 1.4\%.

\noindent\textbf{RoBERTa-Large: }achieves a score of 90\% on MNLI and, as Table~\ref{tbl:Roberta_large} shows, is less sensitive to quantization compared to the RoBERTa. By quantizing the Value and Intermediate layers to 4b for the first 14 Encoders (out of 24) and to 3b for the rest \OURL achieves less than 1\% loss in accuracy.

\begin{table}
\centering
\scriptsize
\caption{HAT model size (M), embedding (E) and compression ratio (CR).}
\label{tbl:HAT}
\ra{1.2}
  \renewcommand{\familydefault}{\sfdefault}\normalfont\setlength{\tabcolsep}{4pt}
\begin{tabular}{c|cccccr}

\textbf{HAT}: WMT'14 En-Fr & \textbf{BLEU}          & \textbf{Error}        & \textbf{M(MB)} & \textbf{E(MB)} & \textbf{M CR }   & \textbf{CR}               \\ \toprule
Baseline FP32 \cite{HAT}              & 41.8          & -            & 227             & 217                 & 1.0$\times$    & 1.0$ \times$             \\
KMeans 4-bit (M) \cite{HAT}             & 41.1          & 0.7          & 28              & 217                 & 8.0$\times$    & 1.8$\times$           \\
\OURL 4 bit (M)             & \textbf{41.5} & \textbf{0.3} & 29              & 217                 & 7.8$\times$ & \textbf{ 1.8$\times$} \\
\OURL 4-bit (E+M)           & 41.4          & 0.4          & 29              & 28                  & 7.8$\times$ & \textbf{7.8}$\mathbf{\times}$ 
\end{tabular}
\vspace{8pt}
\centering
\ra{1.1}
  \renewcommand{\familydefault}{\rmdefault}\normalfont
\scriptsize
\caption{SpanBERT FP16 model quantization.}
\label{tbl:SpanBERT}
\ra{1}
\ra{1.2}
  \renewcommand{\familydefault}{\sfdefault}\normalfont
\begin{tabular}{c|ccccc}
\multirow{2}{*}{\textbf{SpanBERT}} & \multicolumn{2}{c}{\textbf{SQuAD v1.1 }} & \multicolumn{2}{c}{\textbf{SQuAD v2} } & \multirow{2}{*}{\textbf{CR}} \\
                          & \textbf{F1 Score}           & \textbf{Error}           & \textbf{F1 Score}      & \textbf{Error}                &                     \\ \toprule
Baseline FP16             & 93.96\%            & -               & 88.68\%       & -                    & 1.00$\times$           \\
\OURL 3-bit                & 93.15\%            & 0.81\%          & 88.68\%       & 0\%                  & 5.31$\times$        \\
\OURL 4-bit                & 93.94\%            & 0.02\%          & 88.76\%       & \textbf{-0.08\%}     & 3.99$\times$       
\end{tabular}
\vspace{8pt}
\centering
\scriptsize
\ra{1.1}
  \renewcommand{\familydefault}{\rmdefault}\normalfont
\caption{Embedding size (MB) and compression ratio (CR).}
\label{tbl:Embedding_size}
\ra{1}
\ra{1.2}
  \renewcommand{\familydefault}{\sfdefault}\normalfont
\setlength{\tabcolsep}{4pt}
\begin{tabular}{c|c|cccc}
                     & \textbf{Baseline}  & \multicolumn{4}{c}{\textbf{\OURLCORE}}                                                         \\
\textbf{Model/Task}         & \textbf{FP32}      & \textbf{3-bit}    & \multicolumn{1}{c|}{\textbf{CR}} & \textbf{4-bit}    & \textbf{CR} \\ \toprule
BERT-Base/MNLI     & 89.42 MB   & 8.63   & \multicolumn{1}{c|}{10.36$\times$}             & 11.42  & 7.83$\times$              \\
BERT-Large/SQuAD v1.1   & 119.22 MB  & 11.26  & \multicolumn{1}{c|}{10.59$\times$}             & 14.98  & 7.96$\times$              \\
DistilBERT/MNLI    & 89.42 MB   & 8.85   & \multicolumn{1}{c|}{10.10$\times$}             & 11.63  & 7.69$\times$             \\
RoBERTa/MNLI       & 147.26 MB  & 14.18  & \multicolumn{1}{c|}{10.38$\times$}             & 18.77  & 7.84$\times$             \\
RoBERTa-Large/MNLI & 196.34 MB  & 18.41  & \multicolumn{1}{c|}{10.66$\times$}             & 24.55  & 8.00$\times$            
\end{tabular}

\end{table}

\noindent\textbf{HAT: }We evaluated \OURL~on HAT produced models~\cite{HAT}. HAT is a neural architecture search (NAS) method that composes Transformer models for efficient language translation on various hardware platforms. HAT designs FP32 models; however, the authors report accuracy when the model is quantized using KMeans with linear initialization and for a maximum of 300 iterations (similar to Deep-compression~\cite{han2015deep}). Table~\ref{tbl:HAT} shows the BLEU score on the WMT'14 En-Fr, English to French translation task when it is quantized by KMeans (as reported by the authors~\cite{HAT}) and \OURL. \OURL~quantization achieves 0.4 higher BLEU score at the expense of less than 1\% extra footprint for the outliers. If embedding tables are also quantized with \OURL~the compression ratio rises to $7.8\times$. \OURL~proves effective for an attention model (different than the BERT family) where it outperforms a state-of-the-art dictionary-based quantization method.

\noindent\textbf{SpanBERT:} Certain architectures introduced 16b floating-point arithmetic (FP16 or Bfloat16) targeting deep learning training and inference. In general, BERT models are trained with FP32. However, Facebook's SpanBERT is a BERT-derived model that has been successfully trained to work with FP16. Table~\ref{tbl:SpanBERT} evaluates \OURL SpanBERT~\cite{m2019spanbert} on the SQuADv1.1 and SQuADv2 datasets. \OURL with 3b matches the baseline accuracy on SQuADv2 and incurs less than 1\% error in SQuADv1.1. \OURL~achieves a $5.31\times$ compression ratio. This result shows that \OURL remains effective even when using FP16 is possible.


\noindent\textbf{Embedding Table Quantization: }\OURL can also be used to quantize the embedding tables. Table~\ref{tbl:Embedding_size} shows the size of embedding table before and after quantization. The outlier threshold for all of these experiments is set to $\mhyphen 4$. Figure~\ref{fig:EmbedQuant} compares the inference accuracy of each model in two scenarios: in the first, we quantize the embedding layer only and keep the model weights in their original FP32 representation. This experiment illustrates the effect of embedding quantization on the original model's accuracy. This approach not only maintains the model's accuracy but, in certain cases, improves it. In the second scenario, we apply \OURL quantization throughout. While the FC layers' weights are quantized to 3b,
using 4b embedding maintains accuracy, whereas 3b incurs a 0.2\% accuracy loss.

\section{Memory Compression}
\label{sec:memcompression}
\begin{figure*}[t!]
\centering
\begin{minipage}{0.35\textwidth}
\centering
\subfloat[Weight Matrix]{\label{fig:Weight_Matrix}\includegraphics[width=0.65\textwidth]{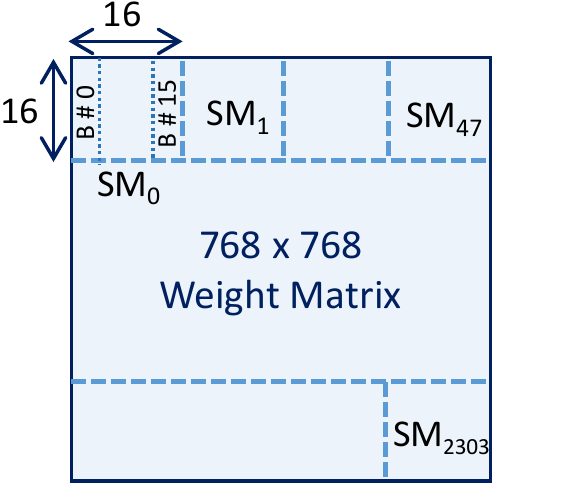}}\\
\subfloat[Effect of SM size on Compression Ratio]{\label{fig:base_compression}\includegraphics[width=0.95\textwidth]{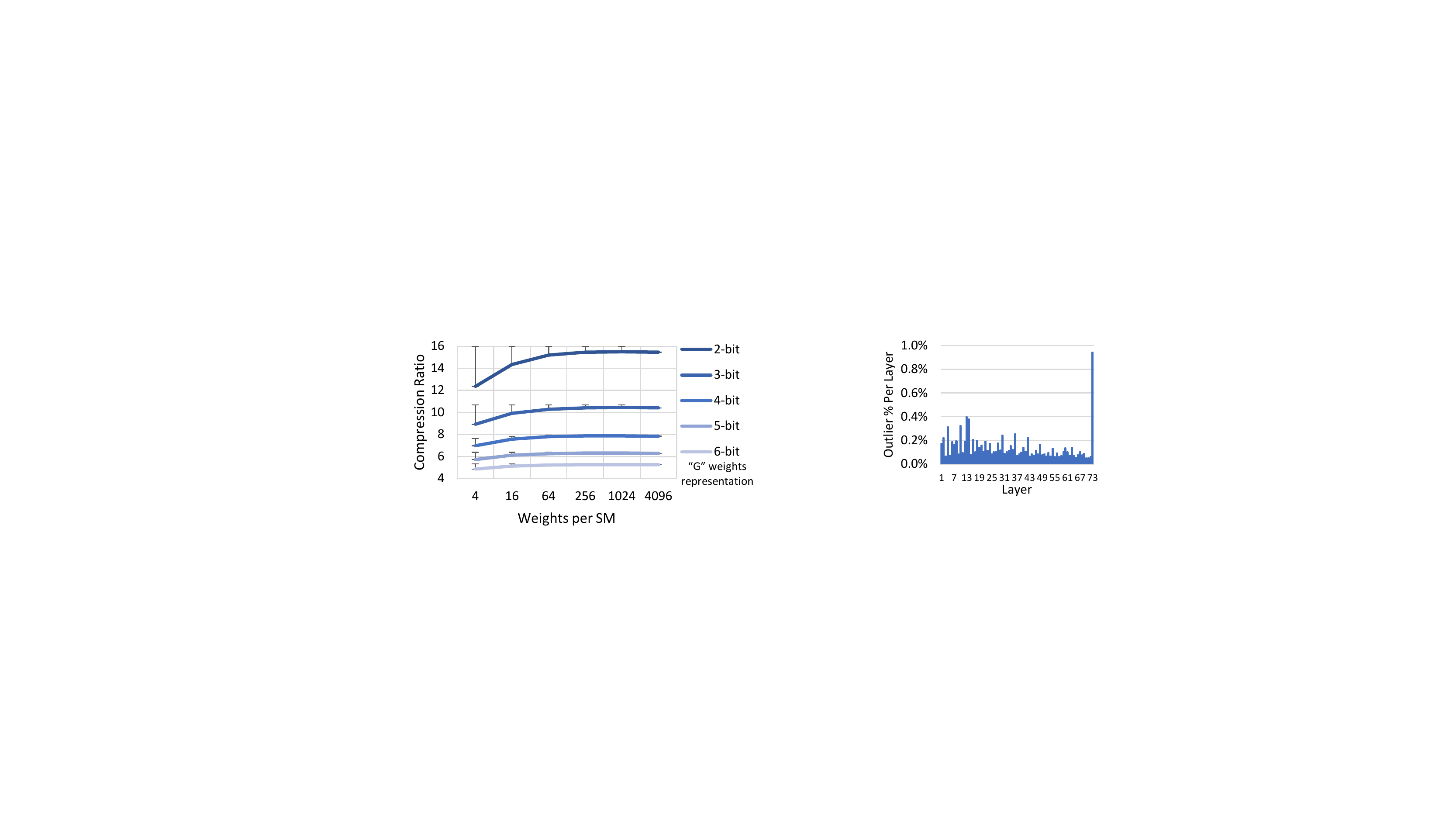}}
\end{minipage}%
\begin{minipage}{0.65\textwidth}
\centering
\subfloat[DRAM Layout and Decompression Engine]{\label{fig:DRAMLAY}\includegraphics[width=0.95\textwidth]{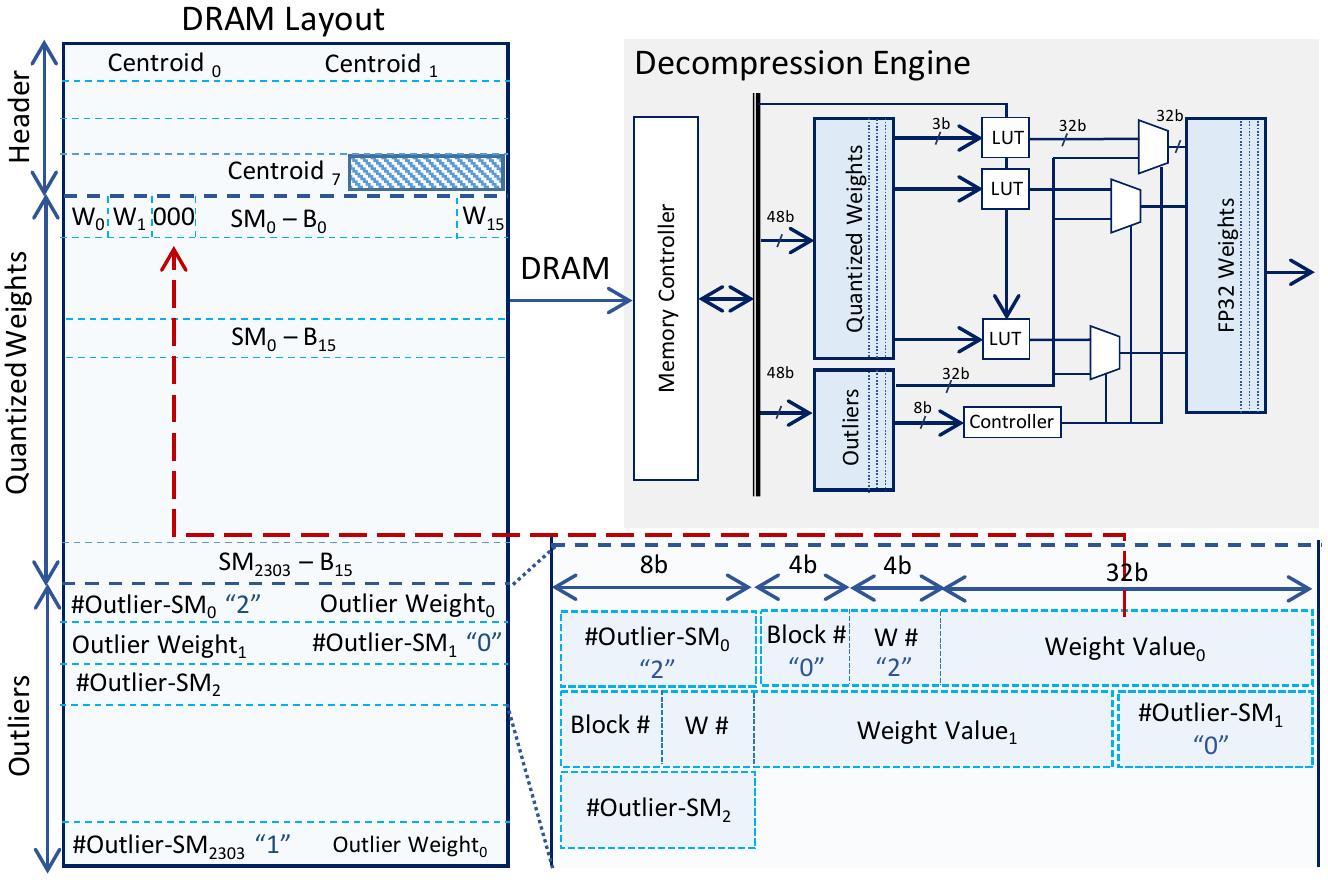}}
\end{minipage}
\caption{\OURL Memory Compression.}
\end{figure*}

This section presents the first practical application of \OURL where it is used to compress weights in off-chip memory. As a result \OURL reduces memory traffic and energy and increases effective memory capacity. 
Regardless, of the dataflow used, the weights of each layer will be accessed sequentially, in large enough chunks to maximize reuse. Accordingly, a format that supports streaming accesses into the weight matrix is sufficient. There is no need to support random accesses. We use the example weight matrix of $768\times768$ of Figure~\ref{fig:Weight_Matrix} to explain how \OURL encodes values in memory and how it decompresses them. For an arbitrary dataflow, assume the weight matrix is divided into $16\times16$ submatrices (SM) of 256 weights each. Each SM is divided into 16 \textit{blocks} where each block contains the weights that are going to be processed together by some data-parallel functional unit, e.g., a GPU shared multiprocessor, and thus need to be stored together in on-chip memory. 

With \OURL, the main challenge is handling outliers. Fortunately, we can still maintain long sequential accesses off-chip thus not sacrificing bandwidth utilization. 
Figure~\ref{fig:DRAMLAY} shows the \OURL \textit{container} comprising three sections: \textit{Header}, \textit{Quantized Weights}, and \textit{Outliers}. The header contains the metadata describing the dimensions of the layer, the number of bits used per weight index, followed by the centroid table. The figure shows only the centroid table and assumes 8 centroids. Padding is used to keep the Quantized Weights section that follows properly aligned to a memory row. This section contains the weight indexes which are stored in exactly the same \textit{order} as the original FP32 weights would have been. However,  \OURL stores just a 3b per weight including the outliers. However, for the outliers the index is ignored and is superseded by the Outliers section. For instance, the third weight in $SM_0-B_0$ is an outlier, and a dummy $"000"$ index is stored for it. This format maintains the relative position of weights and avoids the hardware support and run-time costs of data movement and re-alignment during decompression.

The Outliers section encodes the outliers in submatrix order. Each SM begins with an outlier count (8b supporting up to 256 outliers) followed by the outliers in block order. Each outlier is encoded with a triplet $(B,W,V)$ where $B$ (4b) is the relative block within the SM, $W$ is the weight offset (4b) within the block, and $V$ is the FP32 value. In our example, $SM_0$  contains $2$ outliers 
with the first replacing the third weight in block 0 (dashed arrow).  

To make quantization transparent to the processing elements, the decompression engine generates a stream of FP32 weights. The decompression engine requires two concurrent sequential access streams. The first reads the Header and then the Quantized weights. The Header is used to set the lookup tables (LUT) in the decompression engine. There is one LUT per weight that we want to decompress concurrently, a configuration parameter. Once the header is read, the first stream processes the weight indexes placing them in a FIFO which feeds the LUTs that replace them with the appropriate centroids. Concurrently, the second stream reads in the Outliers section placing values in another FIFO. Using the information per outlier, the outliers selectively overwrite the LUT provided values. 
Since outliers are rare, processing at most one per cycle proves sufficient.

Figure~\ref{fig:base_compression} justifies our choice of $16\times 16$ submatrices. It shows the compression ratio for the BERT-Base model for various SM sizes and for weight quantizations of 2b to 6b. The error bars show what the compression rate would have been if there were no outliers at all. In this case, every weight would be represented by $log_2 (\#bins)$ bits. This representation serves as a useful measure for judging how well the method works. The measurements show that using SMs of 256 weights or more, achieves nearly the maximum possible compression ratio for all weight quantization widths.

The described memory layout is suitable when the dataflow is pre-determined making sequential accesses to all components possible. To allow the dataflow to change without changing the layout, minor changes to the Outlier memory layout are sufficient. The outlier counts and the outliers can be stored separately into two linear arrays $C$ and $O$ respectively. $C$ contains the counts in cumulative form (entry $C_i$ reports the number of outliers before the  \textit{i}th SM) so that they can serve as indexes into $O$.  
To access the outliers of the \textit{i}th SM, an extra memory reference will first read $C_{i}$ and use it as an index to start fetching the outliers from $O$.  The number of outliers contained in the SM is given by $C_{i+1}-C_{i}$. 

\section{Compute Acceleration}
\label{sec:compute}
Most of the computation occurs in the FC layers, each of which is a weight matrix and activation vector multiplication. The use of a very small dictionary allows us to transform the computation from many ``lookup to translate weight index to its FP32 centroid followed by a multiply-accumulate with an FP32 activation'' (one per weight) to an FP32 activation accumulation per weight index followed by a few FP32 multiply-accumulate operations (one per centroid per output activation). That is, rather than first decoding each weight to its centroid and then multiplying with the corresponding activation, we can instead, given a weight index, accumulate the corresponding activation into a per centroid sum, and after doing so for all quantized weights, we can multiply just these \emph{few} sums with the centroids. Consider for example, a layer with 4 input activations $A_i$ and one output activation $OA$.
When weights are not quantized this rudimentary FC layer performs the computation in Eq.~\ref{EQ:FC0}.
\begin{align}
\scriptsize
\label{EQ:FC0}
OA&={{A}_{1}}{{W}_{1}}+{{A}_{2}}{{W}_{2}}+{{A}_{3}}{{W}_{3}}+{{A}_{4}}{{W}_{4}} \\ \label{EQ:FCa}
& ={{A}_{1}}{V_{({W'}_{1})}}+{{A}_{2}}{V_{({W'}_{2})}}+{{A}_{3}}{V_{({W'}_{3})}}+{{A}_{4}}{V_{({W'}_{4})}}\\
 &={{A}_{1}}{V_1}+{{A}_{2}}{V_2}+{{A}_{3}}{V_2}+{{A}_{4}}{V_1}\\
\label{EQ:FCf} &=\underbrace{({{A}_{1}+{A}_{4}})}_a{V_1}+\underbrace{({{A}_{2}}+{{A}_{3}})}_b V_2
 \end{align}
Assume that the weights are quantized to two centroids $V_1$ and $V_2$, with $W_1$ and $W_4$ mapping to $V_1$, and $W_2$ and $W_3$ mapping to $V_2$. If $W'_i$ the 1b weight indexes to the $V$ dictionary values, and $V_{(W'i)}$ refers to mapping a weight index $W'_i$ onto its corresponding centroid, the computation can be transformed as shown in Eqs.~\ref{EQ:FCa}-~\ref{EQ:FCf}. In Eq.~\ref{EQ:FCf}, $a$ is the sum of all activations whose weight maps to centroid $V_1$ and $b$ the sum of all activations whose weight maps to centroid $V_2$. When the number of centroids is much smaller than the number of activations, accumulating the activation values per centroid, and then performing one dictionary lookup and a multiplication per centroid has the potential to reduce energy since FP32 multiplications are more expensive than additions. For  example, for a $768\times768$ FC layer with 3b weight quantization, the first approach would require per output activation 768 weight index to centroid lookups, and 768 MAC operations. \OURL can instead use 8 accumulators and perform 768 activation accumulations in total, followed by 8 centroid lookups and 8 MAC operations ($96\times$ fewer), one per accumulator.

{Re-arranging or changing floating-point arithmetic operations is a concern for hardware and software optimizations as it can affect the outcome. Fortunately, in practice \OURL's approach effectively improves calculation fidelity as it uses 8 separate accumulators. Importantly, each accumulator corresponds to a single weight index. The above arrangement defers multiplying with that weight until the very end, and after all relevant activations have been accumulated. In conventional FP32 hardware, these multiplications occur for all activations and weight indexes before accumulation, increasing the chances that fidelity is lost due to weight magnitude differences. The experiments demonstrate that the \OURL's approach preserves accuracy.}

\begin{figure}[t!]
\centering
\subfloat[Tile]{\label{fig:Ourtile}\includegraphics[width=0.47\textwidth]{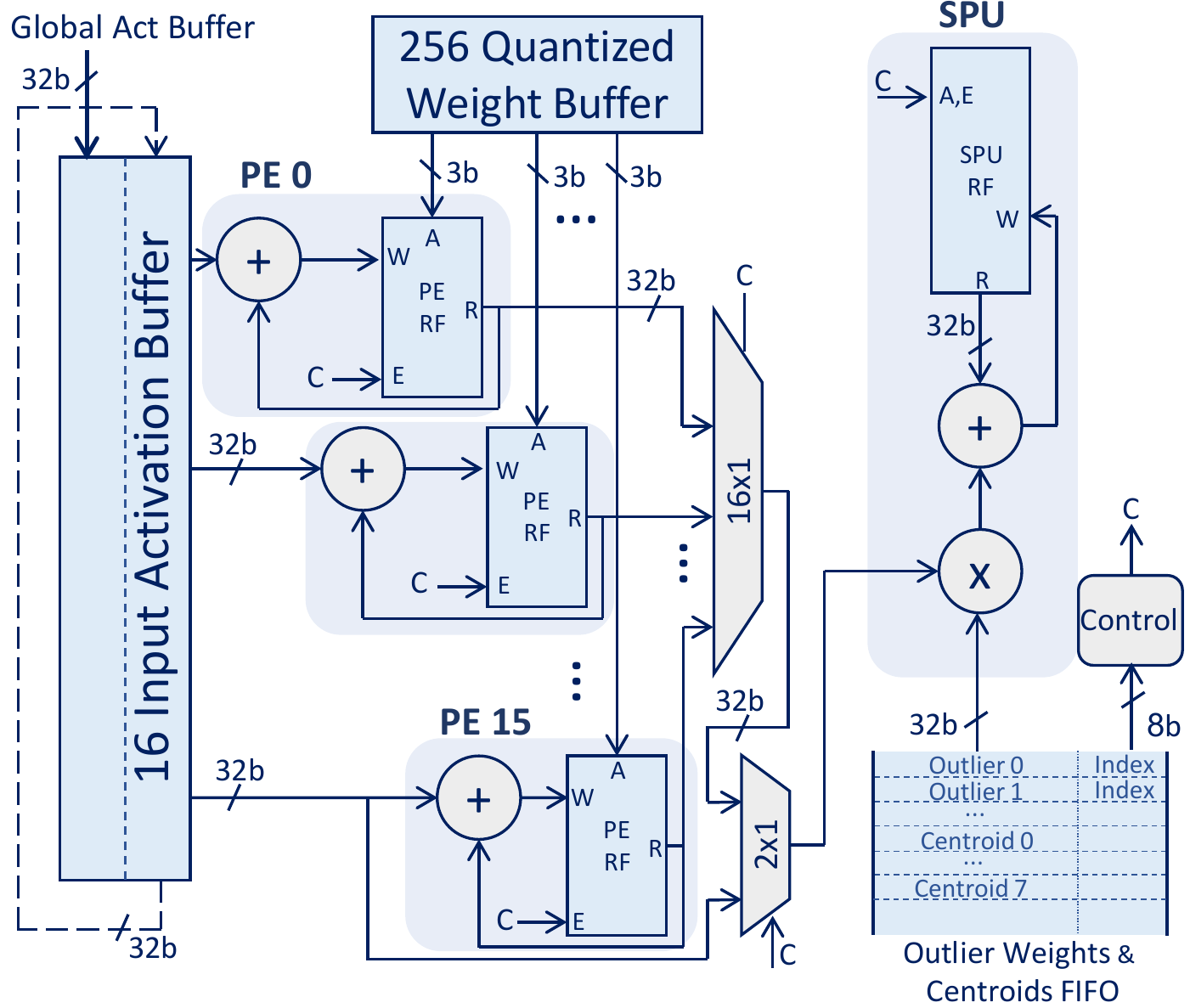}\hspace{10pt} 
}\\
\subfloat[Dataflow]{\label{fig:DF}\includegraphics[width=0.39\textwidth]{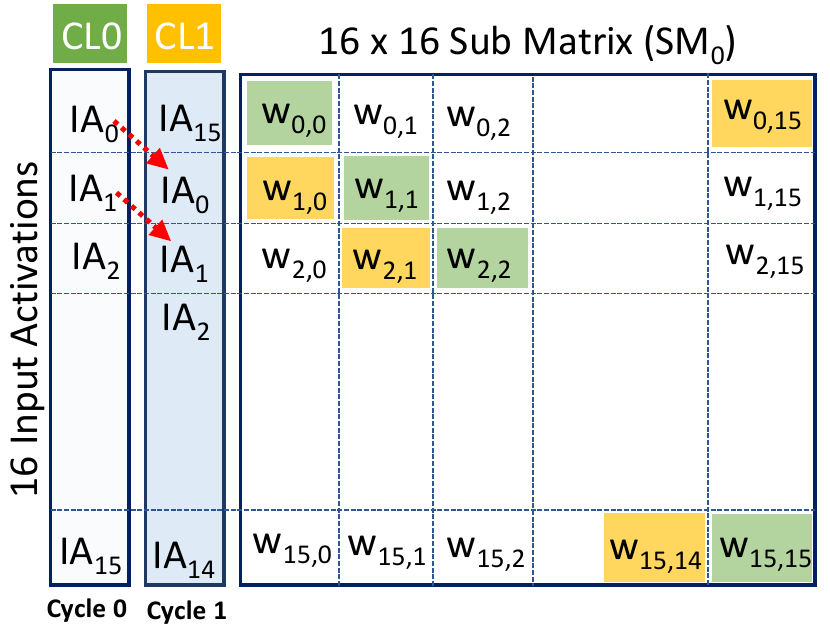}}
\caption{\OURL Hardware Accelerator.}
\end{figure}

\noindent\textbf{Tile Architecture: }
As Fig.~\ref{fig:Ourtile} shows, each \OURL Tile contains a column of 16 Processing Elements (PEs). Each PE contains an FP32 adder and an 8-entry register file, one per possible weight index. The PE is designed to process 3b weight indexes (\OURL can combine tiles to process wider weight indexes). The 16 PEs are all connected to a single Shared Processing Unit (SPU) which has a 16-entry output activation register file, with one entry per PE. The SPU performs the final per centroid MAC operations and also handles the outliers. For this discussion, for clarity, we initially assume an output stationary dataflow where each output activation is produced by one only PE. We have experimented with several dataflows and we describe the one that we use in our experiments later on. Processing proceeds into two phases. In the first phase the PEs accumulate activations in their register files according to incoming ``G'' group weight indexes.  In addition, the SPU is activated any time an outlier is encountered. In the second phase, the SPU processes the sums from each PE, one PE at a time, multiplies them with the centroids, and produces the final output activations.

\noindent\textbf{Phase 1: Per index Activation Accumulation:  }
At the unit's front there are two circular 16-entry activation buffers. At any point, one buffer, the \textit{staging} buffer, loads one activation at a time from the global buffer, while the PEs use the other, the \textit{current} buffer. It takes 16 cycles for the staging buffer to fill, during which the current buffer rotates its entries so that each activation is seen by each PE. Let us ignore outlier processing for the time being. Every cycle, the quantized weight buffer provides 16 weight indexes, one per PE. The PEs use these weights to add the current activation to the corresponding register file entry. Figure~\ref{fig:DF} shows how processing proceeds. For example, in Cycle 0, PE0 uses $W_{0,0}$ to accumulate activation $\mathit{IA}_0$, while PE2 uses $W_{2,2}$ to accumulate activations $\mathit{IA}_2$. In cycle 1, the activations are rotated by one position, and now PE0 uses $W_{0,15}$ to appropriately accumulate activation $\mathit{IA}_{15}$. After 16 cycles, the first block of input activations completes processing, and staging and input buffers switch roles, so that processing can proceed with $\mathit{IA}_{16}$ through $IA_{31}$. The 16 weights along the diagonals in the SM of Figure~\ref{fig:DF} form a block in memory and are stored next to one another as in Figure~\ref{fig:DRAMLAY}. Once all activations have been seen, we proceed to phase two. 

\noindent\textbf{Phase 2: Centroid Processing: }
This phase proceeds in centroid/PE order. As shown in Figure~\ref{fig:Ourtile}, the centroid values are read from the global buffer one by one. Once centroid 0 is read, the SPU uses 16 cycles to multiply it with the corresponding PE register file entry from each PE writing the result into the SPU register file entry for the PE. Then, the SPU proceeds with centroid 1 and so on. In total it takes $8\times16=128$ cycles to finish phase 2. During this time the PEs are idle. Recall that the FC layers are large so at the end this is a small overhead. \OURL will more than make up for this overhead by keeping more weights on-chip and by being able to use more PEs per unit area. If desired, adding more SPUs per tile would reduce this idle time.

\noindent\textbf{Outlier Processing: }
Outliers are processed during phase 1 by the SPU. The SPU reads the outliers from the global buffer into a FIFO. Recall that the outliers are stored using the format of Figure~\ref{fig:DRAMLAY}. Accordingly, the SPU determines the original positions of the outliers, as it encounters them in the order they should be processed according to the described dataflow. For the next outlier in order, the SPU disables accumulation in the corresponding PE when the dummy weight index appear. It then waits until the corresponding activation is rotated to appear at PE15. At that time the SPU can bypass it directly and multiply it with the outlier weight accumulating the result into the appropriate output register file entry (recall that outliers are to be used as-is, they are not mapped to centroids). At the same time PE15 can process the activation as required.  If there are multiple outliers for an activation (the weight matrix has multiple outliers in the same column), the SPU will take as many cycles as needed to process them one by one. During those extra cycles, the PEs are disabled.

Finally, this organization can support 4b weight indexes so that we can execute models such as RoBERTa which require 4b dictionaries for some of their layers. To do so, we pair adjacent tiles so that the 8-entry register files in the first tile store the accumulation of input activations corresponding to the first 8 weight indexes, and the second tile stores the rest. The output register file entries have to be added in pairs to get the final output. This is done over 16 cycles reuse the adder from one of the tiles.

\noindent\textbf{Memory Organization: }
A \OURL accelerator will typically have several tiles. A banked Global Buffer supplies the tiles with the data they need. The most demanding data needs are for the weight indexes. Each tile processes at peak one block of 16 weight indexes per cycle which requires $16\times3b=48b$ per cycle. Each tile also reads a single FP32 activation per cycle at peak. However, the same activation can be shared among all tiles. Finally, each tile also needs to read in outliers which require at most 40b each plus an 8b count per submatrix of 256 weights. However, the outliers are infrequent and the bandwidth needs are low. Accordingly, we partition the Global Buffer across the tiles with at least three banks per tile. One for activations, one wider for weights, and finally one for the outliers plus centroids. Note that there is one set of centroids per layer that we replicate across tiles.

\noindent\textbf{Dataflow: }We have experimented with several dataflows and as previous work have found that the output stationary dataflow is not best~\cite{EyerissISCA2016}. Instead of dedicating each PE to process one output activation at a time, we time multiplex over multiple subsets of activations (from multiple input words) producing partial sums. For this purpose, we block weights in groups of columns, and if beneficial rows, and process the corresponding activation columns producing partial sums. We perform phase 1 and phase 2 processing to produce these partial sums. This allows us to take advantage of weight reuse across words; we use the same weights with the corresponding activations for many words. Once done with all words for one group, we proceed with the next group of columns (and rows) of weights.

\section{Evaluation}
\label{sec:evaluation}
This section demonstrates the performance and energy-efficiency benefits when 1)~\OURL is used to compress data off-chip, and 2)~for an accelerator that uses \OURL compression and the \OURL processing architecture.

We use the models discussed in Section~\ref{sec:models}. The models are quantized with \OURL to 3b except for 12 layers of RoBERTa and 28 layers of RoBERTa-Large that are quantized to 4b as discussed in Section~\ref{Eval:Quant}.
Cycle times are measured using a custom cycle-accurate simulator which uses DRAMsim2~\cite{DRAMsim2} to model off-chip transfers for a DDR4-3200 dual-channel DRAM main memory. The simulator has been tested extensively with microbenchmarks. The simulator computes values faithfully in time which it compares against the expected outputs for correctness. Area and power for the on-chip memories are modeled by CACTI~\cite{CACTI}. Memories are divided into banks to meet the target access time and bandwidth.
We use \textit{post-layout} energy and area estimates: All designs are implemented in Verilog and synthesized using Synopsys' Design Compiler~\cite{design_comp} with a 65nm TSMC technology library and a 1 GHz target frequency. Layouts are generated using Cadence Innovus~\cite{innovus}. For power estimation, we use Intel ModelSim to generate signal activity as input to Innovus. 

\begin{figure*}[t!]
\centering
\captionsetup{font=footnotesize,labelfont=footnotesize}
\begin{minipage}{0.28\textwidth}
\centering
\subfloat[\OURL~Mem. Comp.: Speedup]{\label{fig:performance}\includegraphics[width=0.9\textwidth]{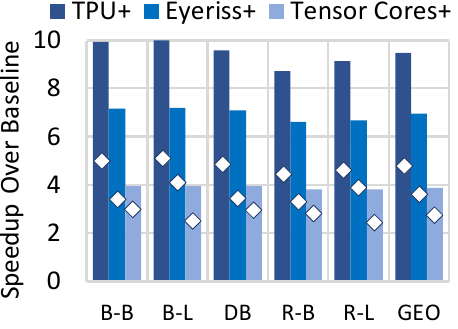}}\\ \vspace{3pt}
\subfloat[\OURL Mem. Comp.: Time Breakdown]{\label{fig:timeBRD}\includegraphics[width=.9\textwidth]{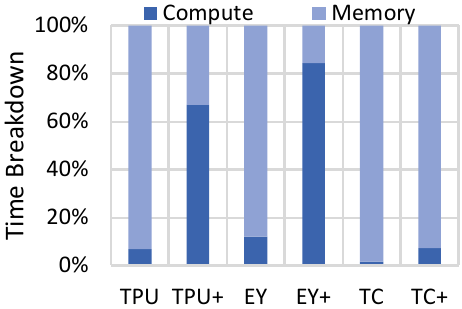}}
\end{minipage}\hspace{-10pt}
\begin{minipage}{0.28\textwidth}
\centering
\subfloat[{\OURL~Mem. Comp.: Relative Energy}]{\label{fig:energy}\includegraphics[width=0.94\textwidth]{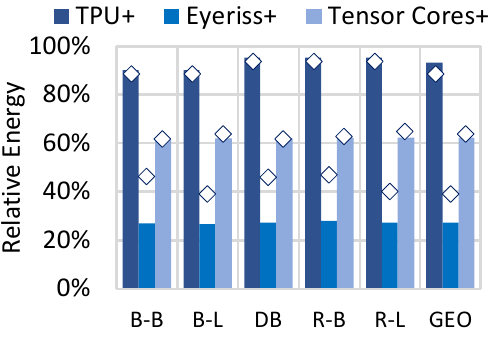}}\\ \vspace{3pt}
\subfloat[\OURL Mem. Comp.: Energy Breakdown]{\label{fig:PowerBRD}\includegraphics[width=.9\textwidth]{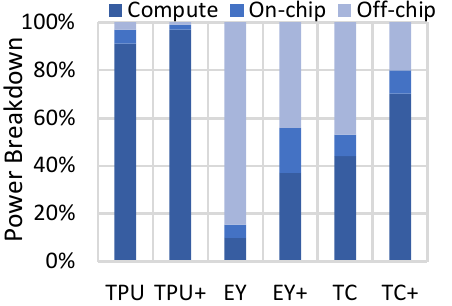}}
\end{minipage}\hspace{-10pt}
\begin{minipage}{0.28\textwidth}
\centering
\subfloat[{\OURL~Accelerator: Speedup over TC.}]{\label{fig:GoboPerf}\includegraphics[width=0.9\textwidth]{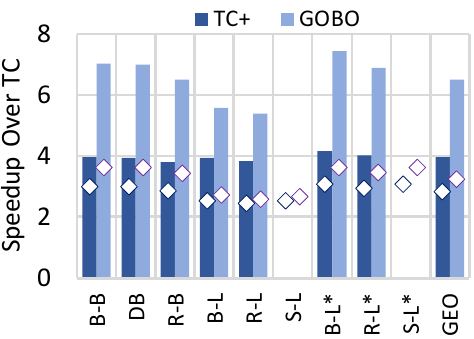}}\\ \vspace{3pt}
\subfloat[{\OURL~Acc.: Relative Energy over TC.}]{\label{fig:GoboEnergy}\includegraphics[width=0.9\textwidth]{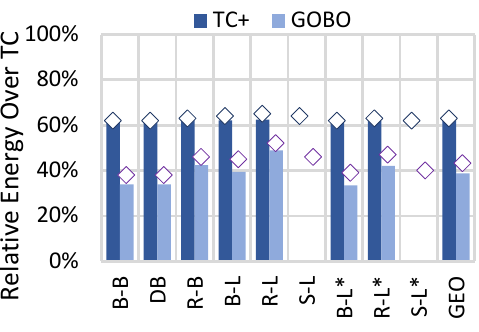}}
\end{minipage}\hspace{-10pt}
\begin{minipage}{0.20\textwidth}
\centering
\subfloat[{Speedup: \OURL Vs. DLP}]{\label{fig:IBMPerf}\includegraphics[width=0.9\textwidth]{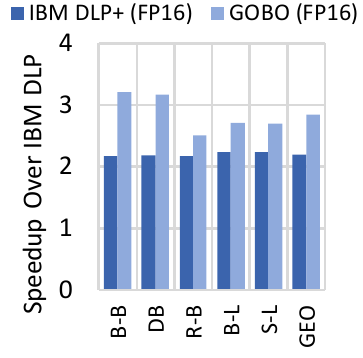}}\\ \vspace{3pt}
\subfloat[{Rel. E.: \OURL Vs. DLP}]{\label{fig:IBMEnergy}\includegraphics[width=0.9\textwidth]{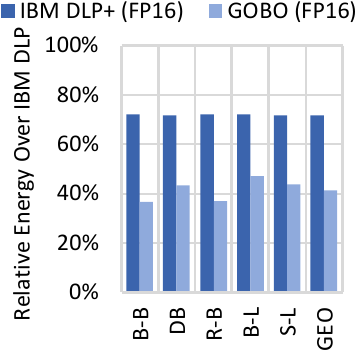}}
\end{minipage}
\captionsetup{font=footnotesize,labelfont=footnotesize}
\caption{Evaluation of \OURL Memory Compression and Hardware Accelerator. {In figures (a) to (f) bars show original FP32 models, and $\mathbf{\diamond}$ markers show scaled FP16 models.} B-B, B-L, DB, R-B, R-L, and S-L are short for BERT-Base, BERT-Large, DistilBERT, RoBERTa-Base, RoBERTa-Large, and SpanBERT-Large respectively. B-L*, R-L*, and S-L* refer to BERT-Large, RoBERTa-Large, and {SpanBERT-Large} models on configurations of TC, TC+ and \OURL with a 4MB on-chip buffer for FP32, and a 2MB for FP16 architectures. EY and TC, stand for Eyeriss and Tensor Cores. TPU+, EY+, TC+, {IBM DLP+} use \OURL off-chip memory compression. {Fine-tuned FP32 version of SpanBERT was not available.}
}
\end{figure*}

\subsection{Memory Compression}
\label{sec:Quant_on_other_arch}
We incorporate \OURL memory compression (Section~\ref{sec:memcompression}) over three popular deep learning hardware architectures. TPU~\cite{TPUISCA17} and Eyeriss~\cite{EyerissISCA2016} are configured as originally presented. To execute these models, however, we implement them with FP32 and FP16 MAC units from a vendor provided library. The MAC units are optimized for energy efficiency. We configure the modeled Tensor Cores accelerator with 128 Tensor Core-like units ~\cite{TC1,TC2} and with a 2MB on-chip buffer. Each Tensor Core unit performs a $A_{4\times4} \times B_{4\times4}$ matrix multiply producing a matrix $C_{4\times4}$ per cycle~\cite{GPGPU}. It is configured to perform 64 MAC operations per cycle. We adjust the dataflow of each architecture to improve performance and energy efficiency.

Figure~\ref{fig:performance} reports speedups when we introduce \OURL memory compression. Speedups are measured over the corresponding architecture without \OURL. Note that these models are not sparse and use FP32 which makes zero compression or other compression methods targeting fixed-point neural networks inapplicable.

\noindent\textbf{TPU: }Performance for the TPU improves nearly by $10\times$. The TPU keeps all activations on-chip and reads weights from off-chip memory. For these networks the TPU is severely memory bound on the weight side thus benefiting fully from the traffic reduction possible with \OURL. Speedups are slightly lower for DistilBERT and the RoBERTa models. The size of the FC layers in DistilBERT leads to different utilization of the TPU's systolic array, and as a result the potential for improvement from memory compression is different than it is for the BERT models. The RoBERTa models have layers that are quantized to 4b instead of 3b. 

\noindent\textbf{Eyeriss: }Speedup on Eyeriss is at nearly $7\times$ on average. Eyeriss optimizes the dataflow to keep MACs busy, and has much fewer MACs than server-class accelerators such as the TPU. Accordingly, the potential for speedup is lower. 

\noindent\textbf{Tensor Cores:} This architecture is also memory bound with $16\%$ of all memory transactions being for activations and partial sums, while weights account for the remaining $84\%$. Given that \OURL reduces weight footprint by nearly $10\times$, performance improves by nearly $4\times$.



\noindent\textbf{Execution Time Breakdown: }Figure~\ref{fig:timeBRD} shows an execution time breakdown for each architecture. The configurations with \OURL memory compression are labeled as TPU+, EY+ (Eyeriss), and TC+ (Tensor Cores) and the graphs report averages over all networks. The measurements confirm that off-chip accesses account for most of the execution time in the baseline architectures.  Using \OURL memory compression, the fraction of time taken by off-chip accesses is reduced to $32\%$ and $15\%$ for TPU+ and Eyeriss+ respectively. Off-chip accesses remain a major bottleneck for TC+ since the activations and the partial sums still require many off-chip accesses. Regardless, TC+ is $4\times$ faster than TC.



\noindent\textbf{Energy: }DRAM transactions are expensive in terms of time and energy. Figure~\ref{fig:energy} reports the energy reduction for each architecture with \OURL memory compression over the baseline. TPU, whose compute grid account for 90\% of overall energ comsumption (Figure~\ref{fig:PowerBRD}), sees the least benefit of 7\%.  {Energy consumption} for Eyeriss improves on average by $3.7\times$ since computation only accounts for 10\% of total energy. In Tensor Cores, off-chip memory accounts for 47\% of total energy, which \OURL reduces to 20\% improving overall {energy consumption} by $1.6\times$ on average. 

\sloppy{
\noindent\textbf{FP16: }Recent architectures add support for 16b floating-point as it has proven sufficient for training and inference of many neural networks. Figure~\ref{fig:performance} ($\diamond$ markers) reports speedups with \OURL~compression assuming the models can use FP16 (presently only demonstrated with SpanBERT). All hardware architectures are configured to use FP16  for off-chip traffic and on-chip processing. \OURL~compression proves effective throughout. Speedups compared to FP32 are halved for the TPU and Eyeriss where weights dominate off-chip traffic. FP16 is less effective for TC as about 16\% of the off-chip traffic is for partial sums. Figure~\ref{fig:energy} ($\diamond$ markers) shows that the relative energy benefits with \OURL compression for the TPU and TC where most energy is consumed in the compute units. The relative benefits scale less for Eyeriss due to its small global buffer (off-chip traffic is more pronounced and using FP16 leaves less potential for compression). Still, even with FP16, \OURL compression reduces energy by $2.5\times$.}



\subsection{\OURLCORE~Hardware Accelerator}
\label{Eval:HW}

We perform an iso-compute-area comparison of the \OURL accelerator with Tensor Cores and IBM DLP~\cite{IBM2018} like architectures. For Tensor Cores we study both FP32 and FP16 variants, and for the IBM DLP we model FP16 hardware as originally proposed. 
We first focus on the FP32 comparison with Tensor Cores. Tensor Cores and \OURL~FP32 designs are given the same 2MB on-chip memory. Table~\ref{tbl:chiparea} summarizes the configurations. The FP32 Tensor Cores' tile is $6.2\times$ larger than the \OURL tile while it has $4\times$ more multipliers and accumulators. Recall that \OURL replaces each multiplier with an accumulator and an 8-entry register file, and that there is only a single MAC unit that is shared across multiple PEs.{ Table~\ref{tbl:TileABD} shows the \OURL tile area breakdown. An FP32 multiplier is $4.1\times$ larger than an 8-entry register file.} As a result, for the same compute area needed by 128 Tensor Cores, we can fit 768 \OURL tiles. 
 
\begin{table}
\centering
\caption{Tensor Cores \& \OURL: Area.}
\label{tbl:chiparea}
\scriptsize
\ra{1.2}
  \renewcommand{\familydefault}{\sfdefault}\normalfont
\setlength{\tabcolsep}{3pt}
\begin{tabular}{cccccc}
             & \multicolumn{2}{c}{\textbf{Tile Area $(mm^2)$}}   & \textbf{Tiles/Chip} & \textbf{MAC/Tile} & \textbf{MAC/Chip}    \\ \toprule
Tensor Cores FP32 & $0.84$& 1$\times$    & 128           & 64       & 1$\times$   \\
\OURL FP32         & $0.13$ & 0.16$\times$ & 768           & 16       & 1.5$\times$ \\ \midrule 
Tensor Cores FP16 & $0.27$ & 1$\times$    & 160           & 64       & 1$\times$   \\
\OURL FP16         & $0.05$ & 0.20$\times$ & 768           & 16       & 1.2$\times$ \\ 
\end{tabular}
\vspace{6pt}
\ra{1.1}
  \renewcommand{\familydefault}{\rmdefault}\normalfont
\caption{\OURL FP32 Tile Area Breakdown}
\label{tbl:TileABD}
\scriptsize
\ra{1.2}
  \renewcommand{\familydefault}{\sfdefault}\normalfont
\begin{tabular}{c|cc|ccc}
\multicolumn{3}{c|}{\textbf{Component Area ($\mu m^2$)}} & \multicolumn{2}{c}{\textbf{Module Area ($\mu m^2$)}} & \textbf{Area/Tile} \\ \toprule
FP32 Adder & 4,149 & 1$\times$ & \multicolumn{1}{c|}{PE} & 6,780 & 80.02\% \\
FP32 Multiplier & 9,379 & 2.26$\times$ & \multicolumn{1}{c|}{SPU} & 17,883 & 13.19\% \\
PE Reg File (8x32) & 2,628 & 0.63$\times$ & \multicolumn{1}{c|}{Buffers} & 8,046 & 5.94\% \\
SPU Reg File (16x32) & 4,630 & 1.12$\times$ &  &  & 
\end{tabular}
\vspace{8pt}
\centering
\ra{1.1}
  \renewcommand{\familydefault}{\rmdefault}\normalfont
\caption{{Tensor Cores} \& \OURL: Absolute Perf. \& Energy.}
\label{tbl:AbsPerf_Power}
\scriptsize
\ra{1.2}
  \renewcommand{\familydefault}{\sfdefault}\normalfont
\begin{tabular}{ccccccc}
                                  & \multicolumn{2}{c}{\multirow{2}{*}{\textbf{Peak TFLOPS}}} & \multicolumn{4}{c}{\textbf{BERT-Base (MNLI)}}                              \\ \cline{4-7} 
                                  & \multicolumn{2}{c}{}                             & \multicolumn{2}{c}{\textbf{Cycles (M)}}   & \multicolumn{2}{c}{\textbf{Energy (J)}} \\ \hline
\multicolumn{1}{c|}{\textbf{Architecture}} & \textbf{FP32}         & \multicolumn{1}{c|}{\textbf{FP16}}         & \textbf{FP32} & \multicolumn{1}{c|}{\textbf{FP16}} & \textbf{FP32}           & \textbf{FP16}          \\ \toprule
TC                                & 16           & 20                                & 70.2 & 35.3                      & 0.24           & 0.11          \\
TC+                               & 16           & 20                                & 17.8 & 11.8                      & 0.15           & 0.07          \\
\OURL                              & 24           & 24                                & 9.9  & 9.7                       & 0.08           & 0.04         
\end{tabular}
\end{table}

\noindent\textbf{Performance: }Figure~\ref{fig:GoboPerf} reports the performance of \OURL relative to the baseline Tensor Cores (TC) and the Tensor Cores with \OURL off-chip compression (TC+).  In this configuration, TC+ is about $4\times$ faster than TC across all models. \OURL is $7\times$ faster than TC for BERT-base and DistilBERT.  Speedups with \OURL are slightly lower for RoBERTa as peak throughput is halved for those layers that are quantized to 4 bits. There are two primary reasons why \OURL improves performance. First, it has a higher peak compute throughput than TC as it can afford to pack more compute units for the same area. Second, it better utilizes the on-chip memory buffers as it keeps weights encoded throughout.  
\OURL's speedup over TC on the large models caps at $5.5\times$ due to size of the on-chip buffer. To show how performance would scale with a larger on chip buffer we evaluated BERT-Large (B-L*) and RoBERTa-Large (R-L*) with TC, TC+ and \OURL configurations using a 4MB on-chip buffer. As expected, this results in higher speedups.

\noindent\textbf{Energy: }Relative energy trends in Figure~\ref{fig:GoboEnergy} follow those of performance closely. For example, \OURL consumes $3\times$ less energy than TC on BERT-base and DistilBERT. Compute accounts for a large portion of TC, so avoiding FP32 multiplications greatly improves energy consumption. Table~\ref{tbl:AbsPerf_Power} reports absolute performance and energy for BERT-Base on TC, TC+ and \OURL. Overall, \OURL is $21\times$ more energy efficient than TC.

\noindent\textbf{FP16:} We scale the FP32 configurations to use FP16 instead. Table~\ref{tbl:chiparea} reports the configurations when both TC and \OURL~use FP16 units and a 1MB on-chip buffer. Given the non-linear scaling of FP multipliers, the ratio of TC to \OURL~tiles decreases from $6.2\times$ to $5\times$. We keep the same number of \OURL tiles and increase the TC tiles so that both use the same compute area. Figure~\ref{fig:GoboPerf} ($\diamond$ markers) shows the relative performance of FP16 TC+ (uses \OURL~compression) and \OURL~over FP16 TC. \OURL~compression makes TC+ $2.8\times$ faster than TC and the \OURL~accelerator is even faster at $3.3\times$. Moreover, \OURL~remains the most energy efficient architecture of the three. Figure~\ref{fig:GoboEnergy} reports that \OURL~consumes $2.4\times$ less energy compared to TC. Using just \OURL~compression improves TC+'s energy over TC by $1.6\times$.  Table~\ref{tbl:AbsPerf_Power} reports absolute performance and energy for these FP16 configurations.

\noindent\textbf{IBM DLP:} IBM's Deep learning Processor~\cite{IBM2018} is an FP16 systolic array with a 2MB on-chip buffer. We demonstrate that a)~IBM DLP benefits from \OURL~memory compression (IBM DLP+), and b)~under iso-compute-area constraints \OURL~accelerator is higher performing and more energy efficient. We configure DLP's systolic array with 512 MACs to match the 1K FLOPS/cycle originally reported. \OURL's tiles are $1.2\times$ smaller than DLP's allowing us to fit 20\% more tiles. \OURL~is configured with the same 2MB on-chip capacity.  Figures~\ref{fig:IBMPerf} and~\ref{fig:IBMEnergy} report \OURL~Accelerator's and IBM DLP+'s performance and relative energy over the IBM DLP baseline.  \OURL~memory compression increases PE utilization from 34\% to 75\% in DLP+, making it $2.2\times$ faster, whereas \OURL~is even faster at $2.85\times$ as its PE utilization is 83\%. \OURL~is the most energy efficient:  $2.4\times$ vs. DLP+ and $6.8\times$ vs the original DLP.

\section{Conclusion}
We presented a post-training quantization method for attention-based NLP models. We demonstrated that it can significantly reduce model size (parameters and embedding tables) with little or no effect on accuracy. We presented two practical applications. In the first, which is plug-in compatible with other accelerators, \OURL is used to compress data from off-chip boosting energy efficiency and off-chip capacity. In the second, we presented an accelerator that uses \OURL quantization throughout. It is unique in that it never expands the weights into their original values directly. 

\section*{Acknowledgement}
This work was supported by the NSERC COHESA Strategic Research Network and an NSERC Discovery Grant. The University of Toronto maintains all rights on the technologies described.



\bibliographystyle{IEEEtran}
\bibliography{IEEEabrv,bibtex/bib/IEEEexample.bib}

\end{document}